\documentclass[journal]{IEEEtran}

\usepackage[protrusion=true,expansion=true,tracking=true]{microtype}
\usepackage[style=USenglish]{csquotes}
\usepackage{textcomp}

\newcommand{\tu}{\textsubscript}

\usepackage{cite}

\usepackage{graphicx}
\usepackage{silence}
\usepackage{amsmath,amssymb,amsfonts}
\usepackage{array}
\usepackage{tabularx}
\usepackage{booktabs}
\usepackage{multirow}
\newcolumntype{Y}{>{\raggedleft\let\newline\\\arraybackslash\hspace{0pt}}X}
\newcolumntype{Z}{>{\centering\let\newline\\\arraybackslash\hspace{0pt}}X}

\usepackage[caption=false]{subfig}

\usepackage{float}

\usepackage{url}
\usepackage{hyperref}
\hypersetup{colorlinks=true,linkcolor=blue,citecolor=blue,filecolor=blue,urlcolor=blue}
\usepackage{enumitem}
\setlist[description]{noitemsep}
\setlist[enumerate]{noitemsep}
\setlist[itemize]{noitemsep,topsep=0pt}

\usepackage[dvipsnames,svgnames,x11names]{xcolor}

\usepackage{tikz}
\usetikzlibrary{calc,positioning,shapes}
\tikzset{
    ncbar angle/.initial=90,
    ncbar/.style={
        to path=(\tikztostart)
        -- ($(\tikztostart)!#1!\pgfkeysvalueof{/tikz/ncbar angle}:(\tikztotarget)$) 
        -- ($(\tikztotarget)!($(\tikztostart)!#1!\pgfkeysvalueof{/tikz/ncbar angle}:(\tikztotarget)$)!\pgfkeysvalueof{/tikz/ncbar angle}:(\tikztostart)$) 
           \tikztonodes
        -- (\tikztotarget) 
    },
    ncbar/.default=0.5cm,
}

\usepackage{lipsum}

\begin{document}

\title{On the Feasibility of Creating Iris Periocular Morphed Images}

\author{
    Juan~E.~Tapia,~\IEEEmembership{Member,~IEEE,}
    Sebastian~Gonzalez,
    Daniel~Benalcazar,~\IEEEmembership{Member,~IEEE,}
    and~Christoph~Busch,~\IEEEmembership{Senior~Member,~IEEE}%
\thanks{Juan Tapia and Christoph Busch are with the da/sec-Biometrics and Internet Security Research Group, Hochschule Darmstadt, Germany, e-mail: \{\href{juan.tapia-farias@h-da.de}{juan.tapia-farias}, \href{christoph.busch@h-da.de}{christoph.busch}\}@h-da.de.}%
\thanks{Sebastian Gonzalez is with the Universidad de Santiago, Santiago, Chile, email: \href{sebastian.gonzalez.s@usach.cl}{sebastian.gonzalez.s@usach.cl}.}%
\thanks{Daniel Benalcazar is a PhD Graduate of Universidad de Chile, Santiago, Chile, email: \href{dbenalcazar@ug.uchile.cl}{dbenalcazar@ug.uchile.cl}.}%
\thanks{Manuscript received Month DD, YYYY; revised Month DD, YYYY.}}

\markboth{Journal of \LaTeX\ Class Files,~Vol.~14, No.~8, August~2015}%
{Shell \MakeLowercase{\textit{et al.}}: Bare Demo of IEEEtran.cls for IEEE Journals}

\maketitle

\begin{abstract}
    In the last few years, face morphing has been shown to be a complex challenge for Face Recognition Systems (FRS). Thus, the evaluation of other biometric modalities such as fingerprint, iris, and others must be explored and evaluated to enhance biometric systems. This work proposes an end-to-end framework to produce iris morphs at the image level, creating morphs from Periocular iris images. This framework considers different stages such as pair subject selection, segmentation, morph creation, and a new iris recognition system. In order to create realistic morphed images, two approaches for subject selection are explored: random selection and similar radius size selection. A vulnerability analysis and a Single Morphing Attack Detection algorithm were also explored. The results show that this approach obtained very realistic images that can confuse conventional iris recognition systems.
\end{abstract}


\IEEEpeerreviewmaketitle

\section{Introduction}
\label{sec:intro}

\IEEEPARstart{T}{he} relevance and opportunity to use the iris as a biometric modality for identification has been increased nowadays, based on the Worldcoin company campaign around the globe\footnote{\url{https://whitepaper.worldcoin.org/}}. This company has captured and enrolled iris images in Near-InfraRed (NIR) from several subjects in many countries to create a token for each person. The token is issued to all network participants to align their incentives around the growth of the network. This could lead the Worldcoin token (WLD) to become the widest-distributed digital asset.

Furthermore, the feasibility of creating double-identity biometric references (different from face) will become a challenge, and has been explored for some EU projects such as iMARS\footnote{\url{https://imars-project.eu/}}; in particular, the biometric characteristics promoted by ICAO 9303. 


Previous research has investigated the feasibility of fingerprint morphing.~\cite{makrushin2021feasibility, ferrara2016feasibility}. These implementations have proven the weaknesses of commercial fingerprint verification systems against morphed samples by developing a new algorithm that is able to generate quite realistic morphed fingerprints.

For the iris modality, some studies have proven the feasibility of creating morphed iris templates resembling information from two iris-codes and textures~\cite{rathgeb2017feasibility, gomez2018predicting}. Proposed techniques have been used in conjunction with template reconstruction methods in order to obtain morphed iris textures that can be used to launch presentation attacks during enrolment and verification. Moreover, adversarial learning techniques were also explored to create realistic iris images~\cite{ogino2024outsmarting}.

To understand this work, we take into account some assumptions for two possible attack scenarios. One, the \enquote{manipulation of a biometric database} scenario, which is the attack scenario that assumes an attacker can access the system's database where iris codes of legitimate subjects are stored. In addition, the attacker knows the software components the system uses to extract iris codes. In the second scenario, \enquote{presentation attack enrolment}, learning about the software components used by the system to extract iris codes is required. However, contrary to the attack outlined in the first scenario, no access to the system database is required. Such an attack might also be feasible using a printed contact lenses in a supervised enrolment scenario. From that point onwards, both subjects can gain access to the system (or share a single electronic travel document). A morphing technique for iris images might also be employed in this type of attack, which would not require a reconstruction of an image from the morphed iris code.

This work explores the morph iris modality from periocular images, using the approach based on landmark points implemented for face-morphing images as a guide, translated into NIR morph periocular iris images. We believe that creating morph periocular images instead of iris codes has the potential to generate all the traditional stages of iris recognition, not only the final codified iris image.

The contributions of this work are as follows:

\begin{itemize}
    \item \textit{State of the Art}: A comprehensive analysis of the state of the art is performed regarding this new modality of iris morphing, encompassing segmentation and iris recognition systems.
    \item \textit{Iris Recognition}: A new iris recognition system is proposed based on a Siamese network architecture.
    \item \textit{Iris Morphing}: Image-level periocular iris morphing modality is proposed, highlighting its feasibility and limitations.
    \item \textit{Pairs selection}: Two-subject pair selection methodologies are proposed to improve the results of iris morphing.
    \item \textit{Vulnerabilities}: Our work shows that iris recognition systems are very sensitive to periocular iris morphed images, and these systems can be attacked with a high chance of success.
\end{itemize}

The rest of the article is organised as follows: Section~\ref{sec:related} summarises the related works on iris morphed and its stages. A new iris recognition system and pair subject selection methods are described in Section~\ref{sec:method}. This work's experimental framework and results are then presented in Section~\ref{sec:experiments}. We conclude the article in Section~\ref{sec:conclusions}.

\section{Related Work}
\label{sec:related}

\subsection{Iris Morphing}

The human iris has a distinctive texture that is ideal for verification and identification, especially in the Near-Infrared (NIR) spectrum~\cite{bowyer2008image}. For the creation of a double identity iris, there are two approaches that can be used: a feature-level approach and an image-level approach.

The feature-level morphing is similar to the GAN-generated double identity fingerprints mentioned previously, where the biometric information is translated into a vector, and the vectors of the two identities are mixed.

Rathgeb et al.~\cite{rathgeb2017feasibility} propose a stability-based bit substitution (SBS), instead of the random substitution of the code bits or the code rows. From the obtained iris code, the new iris image is generated using techniques like in~\cite{galbally2013iris} or~\cite{venugopalan2011generate}.

Gomez-Barrero et al.~\cite{gomez2018predicting} proposed morphing at the feature level, where iris codes are morphed using stability-based bit substitution. The proposed scheme is applied to iris codes extracted from iris images of the CASIAv4-Interval database.

Shechtman et al.~\cite{shechtman2010regenerative} proposed morphing as an optimization problem, in order to achieve bidirectional similarity of each morphed image with its neighbouring frames within the morph sequences as well as the input images. The proposed scheme was applied to CASIA-V3\footnote{\url{http://www.cbsr.ia.ac.cn/english/IrisDatabase.asp}} and IITD-v1~\cite{kumar2010comparison}.

The process for image-level morphing is also familiar~\cite{sharma2021image}. It consists of finding \enquote{corresponding} irises (to create a viable morphed iris, the two original ones need to be relatively similar), warping the images according to landmarks on both images (surrounding the pupil and the iris itself) to align them, and finally blending the two images using linear blending between the pixels of both images.



Venugopalan et al.~\cite{venugopalan2011generate} explored a method of creating iris textures for a given person embedded in their natural iris texture (or someone else’s if desired) using just the iris code of the person. Suppose these textures are used in an iris recognition system (IRS), in that case, they will give a response similar to the original iris texture, showing that presenting an attack texture to an IRS will generate the same score response as the original iris texture.

Renu and Ross~\cite{sharma2021image} proposed morphing two iris images at the image level and analysed the approach with iris images from two different datasets, IITD~\cite{kumar2010comparison} and WVU multi-modal datasets~\cite{crihalmeanu2007protocol}. Commonly, morphing techniques at the image level are landmark-based. They demonstrate the vulnerability of iris recognition methods to morph attacks. Like in morphed fingerprints, the best results appear to be obtained using the image level approach, again, probably due to the information loss in the feature level creation of a new iris from the generated iris-code, while the image level iris would maintain more of these key features. No selection pair subject methods were explored.

\subsection{Segmentation}

Cutting-edge algorithms, like semantic segmentation, are primarily designed to identify complex objects in urban settings, such as cars, buildings, and people, as well as in biometric gaze applications. Several works have been previously developed to segment NIR iris images under challenging conditions using semantic segmentation~\cite{tapia2021semantic, valenzuela2020towards}.

Valenzuela et al.~\cite{valenzuela2020towards} proposed a lightweight and efficient network based on DenseNet56, with fewer parameters than traditional architectures, in order to be used for mobile device applications. As a result, DenseNet10 with only three blocks and ten layers was proposed. The sclera was identified as the more challenging structure to be segmented.

In our previous implementation~\cite{tapia2021semantic}, we proposed a method to develop an efficient framework to locate and segment the iris and pupil in multiple frames for subjects under the influence of alcohol~\cite{valenzuela2020towards}, which was developed to be efficient in the number of parameters and model sizes. This method is more aligned with the task of segmenting morphed images, which present challenging images because of artefacts created from our morphed method with random and pupil-size images. This segmentation method used a mixture of pupil and iris estimators.

\subsection{Iris Recognition}

John Daugman proposed to represent this complex texture in a compact iris code that is easy to compare from subject to subject~\cite{daugman2009iris}. Over time, iris recognition methods have improved using both image processing and deep learning techniques. Nowadays, iris recognition is one of the most reliable biometric methods.

The process of iris recognition consists of image acquisition, segmentation, localisation, normalisation, feature extraction and comparison~\cite{daugman2009iris, bowyer2016handbook}. The feature extractor in traditional iris recognition systems is based on hand-crafted filters, such as Gabor wavelets~\cite{daugman2009iris}, Localized Binary Patterns (LBP)~\cite{he2009efficient}, Binarized Statistical Image Features (BSIF)~\cite{raja2014binarized, rathgeb2016efficient, tapia2019gender}, and 3D descriptors~\cite{benalcazar2020scanner, benalcazar2020recognition}. Those methods produce strong features that help in the subject identification task. However, a question arises on whether they are the best possible filters that could be obtained or not~\cite{czajka2019domain}.

On the other hand, more modern iris recognition systems~\cite{gangwar2016deepirisnet, nguyen2018iris, zhao2017towards, zambrano2022iris} rely on features extracted using CNN pre-trained on ImageNet~\cite{deng2009ImageNet}. Those features are also strong since the network had observed numerous complex patterns in the input images that help classify them in a thousand different classes~\cite{he2016deep}. Based on the previous statements, the strong features produced by pre-trained networks can be successfully used for different tasks, such as iris recognition, even without fine-tuning~\cite{zambrano2022iris}.

After the great success of deep learning for image classification, this technique was quickly adapted to iris recognition. Gangwar and Yoshi developed DeepIrisNet~\cite{gangwar2016deepirisnet}, a CNN that encodes the iris texture robustly and accurately, even for cross-sensor identification.

Zhao and Kumar proposed UniNet, a two-path network that generates the encoding of the iris texture as well as a binary mask that predicts the probability of each element in the feature vector coming from iris and non-iris regions~\cite{zhao2017towards}.

Wang and Kumar considered residual networks with dilated convolutional kernels to optimize the training process~\cite{wang2019toward}. Also, Zhao et al. implemented a method based on capsule network architecture with great performance~\cite{zhao2019deep}.

Minaee et al.~\cite{minaee2019deepiris} developed DeepIris, based on residual layers and used the entire iris image instead of the rubber-sheet model. Adamovic et al.~\cite{minaee2019deepiris} combined stylometric feature extraction with machine learning techniques to produce a novel approach. Additionally, images produced by generative adversarial networks have been used to enhance the training dataset~\cite{adamovic2020efficient, lee2019conditional}.

Boyd et al. compared the performance of iris-specific trained ResNet-50 models against that of ResNet-50 trained for non-iris tasks~\cite{boyd2019deep}. They discovered that fine-tuning a model for iris recognition obtained greater accuracy than off-the-shelf models and training from scratch~\cite{boyd2019deep}.

Additionally, siamese networks~\cite{koch2015siamese} have been trained to obtain specific filters for biometric tests since the one-shot-learning paradigm is very similar to the identity verification task. Zhang et al. developed two Siamese network architectures for face recognition, which were trained from scratch using face images~\cite{sun2021updatable}. 


Ridha and Saud started using Siamese networks for iris recognition~\cite{zhong2018palmprint}. They also implemented a learnable mutual-component distance that allowed their method to obtain good performance even under heterogeneous device conditions ~\cite{zhong2018palmprint}.

Parzianello and Czajka developed a contact-lens-aware iris-recognition method using siamese networks~\cite{parzianello2022saliency}.

One of the last state-of-the-art methods was proposed by Zambrano et al.~\cite{zambrano2022iris}. They proposed a fast iris recognition method that requires a single comparison operation and is based on pre-trained image classification models as feature extractors. Their approach uses the filters of the first layers from CNN as feature extractors and does not require fine-tuning for new datasets.

The challenge with this approach is that ImageNet CNNs have been trained to produce similar features for objects of the same class, such as eyes. There is no emphasis on producing distinctive features for each iris. Choosing earlier layers and a fine-tuning process can certainly increase the network's performance. Another challenge is the scarcity of iris recognition models openly available for research to improve or compare against and the lower number of subject IDs per dataset.

Motivated by previous approaches, we propose an iris recognition method consisting of a siamese network and trained with the triple-hard-loss function. It takes two periocular iris images as the input and outputs the similarity using the Euclidean distance between the embeddings produced by the CNN. 

\section{Metrics}

For this work, several metrics have been applied to measure the similarity and dissimilarity of mated and non-mated samples. The Morph Attack Classification Error Rate evaluates the success of a morphing attack detector, and a vulnerability analysis measures the impact on the new morph images in the iris recognition system.

\subsection{d' Value}

The d-prime ($d'$) metric assesses the separation between the mated and non-mated dissimilarity score distributions~\cite{daugman2009iris}. The higher the value, the better. It is computed using equation~\eqref{eq:d_prime}, where $\mu_m$ and $\mu_n$ are the means of the mated and non-mated, respectively, and $\sigma_m$ and $\sigma_n$ are the standard deviations.

\begin{equation}\label{eq:d_prime}
    d' = \frac{|\mu_m - \mu_n|}{\sqrt{0.5 \times (\sigma_m^2 + \sigma_n^2)}}
\end{equation}

In order to evaluate the biometric performance of our approach, the False Match Rate (FMR) and the False Non-Match Rate (FNMR) metrics are computed. The False Match Rate (FMR) is the number of comparisons falsely classified as mated, divided by the number of non-mated comparisons. Likewise, the False Non-Match Rate (FNMR) is the number of comparisons falsely classified as non-mated, divided by the number of mated comparisons. Both metrics, in tandem, give the performance of the system for a given operating point. 

In this work, three metrics are considered: Equal Error Rate (EER), and the biometric performance at two operational points, FNMR\tu{10}, and FNMR\tu{20}. The EER is the point at which FMR is equal to FNMR. On the other hand, FNMR\tu{10}, and FNMR\tu{20} are the points in which FMR is at 10\%, and 5\% respectively. All those operating points can be illustrated in the Detection Error Trade-off (DET) curve.

\subsection{Morphing Attack Classification Error Rate}

The detection performance of biometric Morphing Attack Detection (MAD) algorithms is standardised by the ISO/IEC DIS 20059~\cite{isoiec20059}. The most relevant metrics for this study are the Morphing Attack Classification Error Rate (MACER) and the Bona fide Presentation Classification Error Rate (BPCER).

The MACER metric measures the proportion of morphing attacks incorrectly classified as bona fide presentations in a specific scenario. The BPCER measures the proportion of bona fide presentations incorrectly classified as morphing attacks. The computation method is detailed in Equation~\ref{eq:apcer}, where the value of $N_M$ corresponds to the number of morphing presentation images, $Res_{i}$ is $1$ if the $i$th image is classified as morphed, or $0$ if it was classified as a bona fide presentation.

\begin{equation}\label{eq:apcer}
    MACER = \frac {\sum_{i=1}^{N_M} {1 - Res_{i}}}{N_M}
\end{equation}

On the other hand, the BPCER metric measures the proportion of bona fide presentations wrongly classified as a morphing attack. BPCER can be computed using Equation~\ref{eq:bpcer}, where $N_{BF}$ is the amount of bona fide presentation images, and $RES_{i}$ takes the same values described in the MACER metric. Together, the two metrics determine the performance of the system, and they are subject to a specific operation point.

\begin{equation}\label{eq:bpcer}
    BPCER = \frac{\sum_{i=1}^{N_{BF}} RES_{i}}{N_{BF}}
\end{equation}


The detection performance at various operation points can be illustrated in a Detection Error Trade-off (DET) curve, which is also reported for all the experiments. 



\subsection{Morph Vulnerability Analysis}

We used as input the morphed iris images for the two iris recognition systems and measured their vulnerability in terms of the Morphing Attack potential~(MAP) ~\cite{isoiec20059}. The MAP is the ratio of successful morph attacks to the total number of morph attacks. A morph attack succeeds when the image matches both component identities at a specified threshold.

The MAP can be estimated for a defined number of attempts with probe samples and for a defined number of iris recognition algorithms. The MAP matrix can report the attack potential by analysing many attempts and from one-to-many recognition systems, where more attempts and more recognition systems represent a stricter measure for vulnerability reporting.



An iris recognition system with high recognition accuracy (i.e. biometric performance) can also have a high vulnerability to iris morphing attacks. As the vulnerability of the IRS can also be impacted by the recognition accuracy, a joint metric---Relative Morph Match Rate (RMMR), was proposed earlier~\cite{scherhag2017biometric}. It is, therefore, essential to first evaluate the biometric performance of the IRS according to international standard ISO/IEC 19795-1~\cite{isoiec19795-1}, and then evaluate its vulnerability by using the pre-set threshold (e.g., $FMR = 0.1\%$), to eventually combine the two measures.

The real strength of an iris recognition system cannot be established unless a good iris recognition performance and robustness with respect to morphing attacks is achieved. 


\section{Dataset}

In this work, we use two datasets. The first one is the Notre Dame Dataset\footnote{\url{https://cvrl.nd.edu/projects/data/\#nd-crosssensor-iris-2012-data-set}}, captured with a Near-Infrarred Range (NIR) LG-4000 sensor, as described in~\cite{benalcazar2023toward}. The composers of the dataset separated the left and right eyes of each subject as different instances (IDs); thus, the dataset was named ND-LG4000-LR. There are 811 IDs and 10,959 images, comprising 5,476 left-eye and 5,483 right-eye images.

\begin{figure}[!htb]
    \centering
    \subfloat{\includegraphics[scale=0.30]{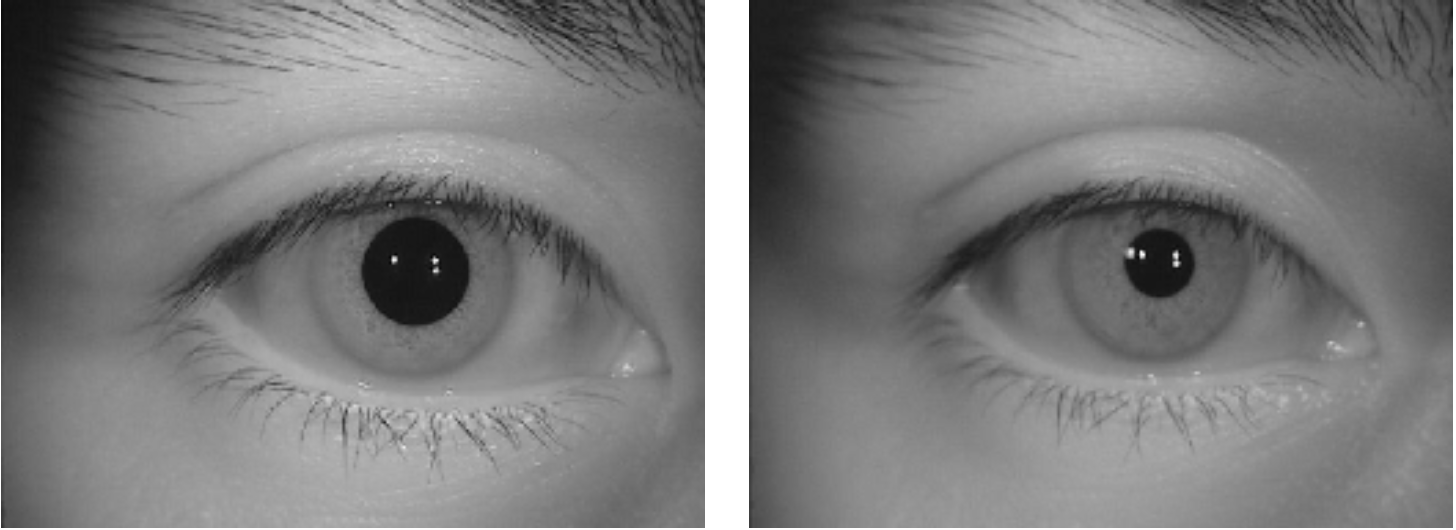}}\\
    \subfloat{\includegraphics[scale=0.17]{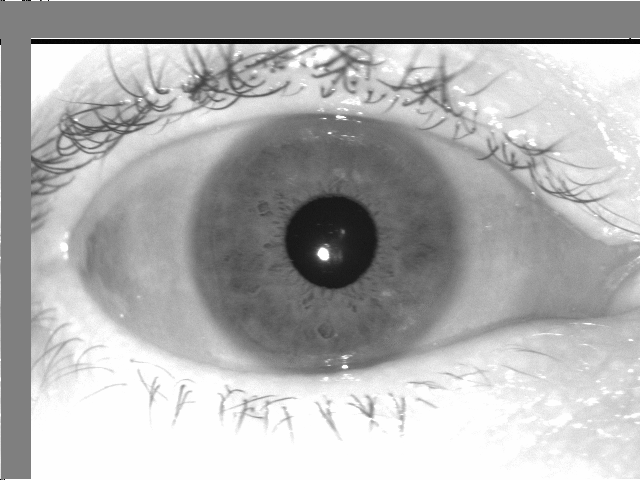}}\hspace{0.7em}%
    \subfloat{\includegraphics[scale=0.17]{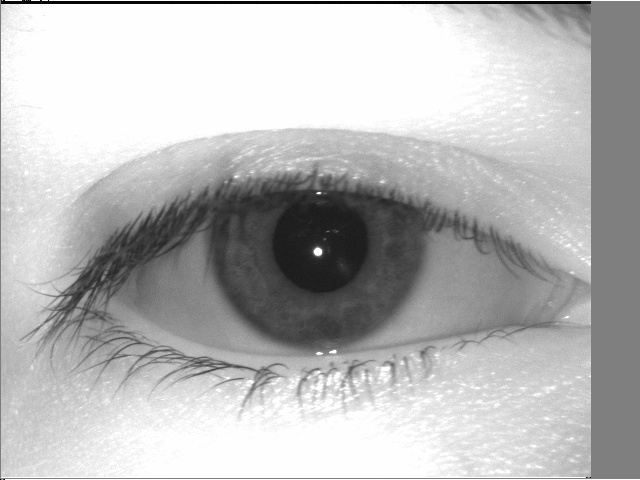}}
    \caption{Top: Example image from CASIA-IrisV4-Lamp. Bottom: UND databases. The grey border is an artefact produced by the capture device.}
    \label{fig:example-database}
\end{figure}

The second one is the CASIA-IrisV4-Lamp, which was collected using a hand-held iris sensor produced by OKI\footnote{\url{https://hycasia.github.io/dataset/casia-irisv4/}}. A lamp was turned on or off close to the subject to introduce more intra-class variations in the CASIA-Iris-Lamp version. Elastic deformation of iris texture, due to pupil expansion and contraction under different illumination conditions, is one of the most common and challenging issues in iris recognition. CASIA-IrisV4-Lamp database contains the left and right eyes for 411 classes, and the size of each image is $480\times 640$. The illumination variations are introduced in this database.  Figure~\ref{fig:example-database} shows examples from both datasets.


\section{Methods}
\label{sec:method}

\subsection{Iris Recognition}

\begin{figure*}[!htb]
    \centering
    \subfloat{\includegraphics[width=0.38\linewidth]{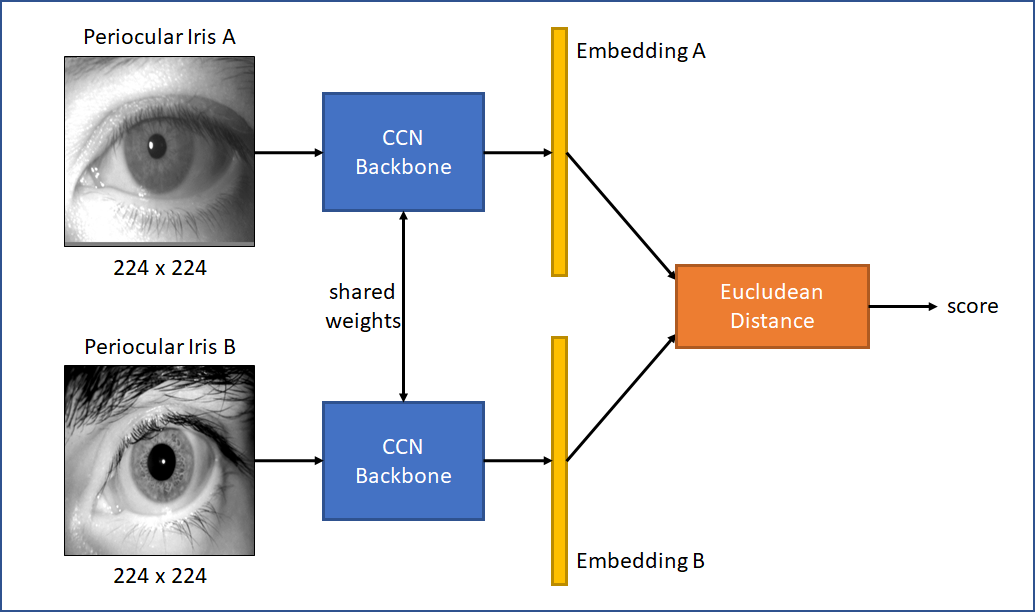}}\hspace{1em}%
    \subfloat{\includegraphics[width=0.38\linewidth]{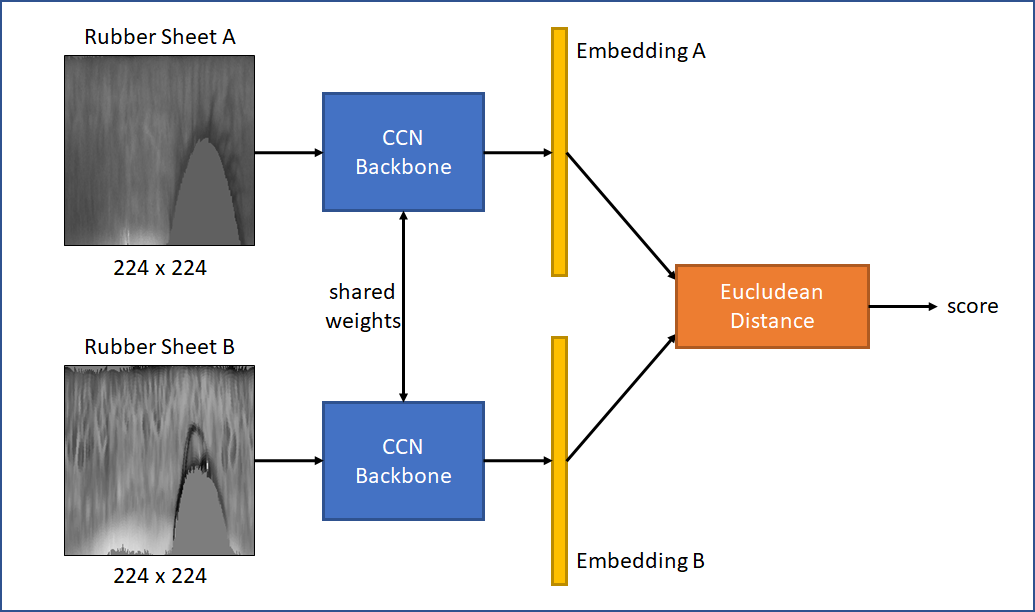}}
    \caption{SiamIris Architecture. This approach predicts whether two irises come from the same or different subjects. This iris recognition system can be processed as periocular images or rubber-sheet images. For example a) uses periocular iris images as inputs, whereas in b) rubber-sheet images are used as inputs.}
    \label{fig:siam}
\end{figure*}

Starting from these constraints, we propose a new iris recognition system based on deep learning and a Siamese network approach, called \enquote{SiamIris}, as is shown in Figure~\ref{fig:siam}. This open-source network uses a Siamese architecture to train CNNs for the specific purpose of iris recognition. It is important to highlight that implementing a new iris recognition system is the starting point for analysing the influence of iris morph images.

As illustrated in Figure~\ref{fig:siam}, our method consists of a Siamese network. It takes two periocular iris images as the input (or two rubber-sheet), and outputs the similarity using the Euclidean distance between the embeddings produced by the CNN. The Euclidean distance is preferred over other distance metrics, such as cosine distance, Mahalanobis, or Manhattan, because the Euclidean distance is easier to differentiate, which produces an easier-to-implement and faster training~\cite{schroff2015facenet}.


Figure~\ref{fig:siam} depicts the network architecture as the two images passing through two identical copies of the CNN block, but in practice, only one copy is stored in memory, and it takes the two images, one at a time, producing the two embeddings. The embeddings \emph{A} and \emph{B} are essentially two feature vectors of size $1\times N$. The value of $N$ depends on the final layer of the backbone network as explained below.

Table~\ref{tab:dataset} indicates the number of instance IDs, images, and comparisons in the test, train, and validation splits. The splits were constructed as an ID-disjoint dataset. Therefore, all of the images of one ID can only be found on either the test, train or validation partitions. Table~\ref{tab:dataset} indicates the dataset has only 487 IDs for training, which is very challenging. 

\begin{table}[!htb]\def\tabularxcolumn#1{m{#1}}
    \centering
    \caption{Description of the ND-LG4000-LR Dataset}
    \label{tab:dataset}
    \begin{tabularx}{\linewidth}{lYYY}
        \toprule
        \textbf{Description}  & \textbf{Test} & \textbf{Train} & \textbf{Validation} \\
        \midrule
        Instances (IDs)       &           162 &            487 &          162 \\
        Periocular Images     &         2,090 &          6,663 &        2,203 \\
        Rubber-sheets         &         2,090 &          6,663 &        2,203 \\
        Mated Comparisons     &        19,029 &         66,114 &       22,109 \\
        Non-mated Comparisons &       105,470 &      1,079,445 &      117,367 \\
        \bottomrule
    \end{tabularx}
\end{table}

For the Siamese network training process, the CNN backbone block used two pre-trained ResNet50~\cite{he2016deep} and MobileNetV2 \cite{howard2017mobilenets} CNNs. Additionally, we compare the use of periocular iris images (Figure~\ref{fig:rubb}\textcolor{blue}{a}) against rubber-sheet images (Figure~\ref{fig:rubb}\textcolor{blue}{e}).

All the networks used the Hard Triplet Loss~\cite{schroff2015facenet} function during training. For the computation of the Triplet Loss ($L_t$), three images are involved: an anchor image ($a$), an image with the same ID as the anchor ($p$), and one image with a different ID ($n$). In this way, the network learns to minimise the distance between images of the same ID while maximising the distance between images of different IDs. Triplet mining is performed within the batch for better training efficiency. For distance $(d)$ on the embedding space, the loss of a triplet $(a,p,n)$ is defined in Equation~\ref{Lt}.

\begin{equation}\label{Lt}
    L_t = \max(d(a, p) - d(a, n) + margin, 0)
\end{equation}

\subsection{Segmentation}

From the segmentation stage based on DenseNet10, we can obtain the radii of the pupil and iris with the actual algorithms included in the implementation based on Least Mean Square (LMS). This method processes the segmentation masks and gives pupil, and iris coordinates as an ellipse [$X$, $Y$, $r_{min}$, $r_{max}$], where $X$ and $Y$ are the centres of the pupil/iris, and the other coordinates are the min-max radii of the ellipse. These morph images are very challenging because, many times, the pupils or the iris are not perfectly aligned, confusing the iris localization with shadows and artefacts.

\begin{figure}[!htb]
    \centering
    \includegraphics[width=0.75\linewidth]{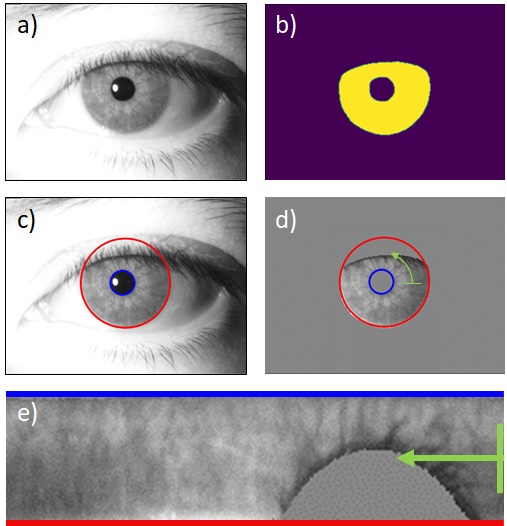}
    \caption{Image pre-processing steps. a) Input image. b) Iris mask predicted by DenseNet10. c) Iris and pupil localization using the mixed algorithm. d) Isolated iris. e) Rubber-sheet model. Blue lines represent the pupil-iris boundary, while red lines represent the iris-sclera boundary. The green arrows in d) and e) represent the direction in which the iris was unwrapped.}
    \label{fig:rubb}
\end{figure}

First, a periocular iris image (Figure~\ref{fig:rubb}\textcolor{blue}{a}) is segmented based on the DenseNet10 network~\cite{valenzuela2020towards}. The result is a semantic segmentation mask of the valid pixels of the iris (Figure~\ref{fig:rubb}\textcolor{blue}{b}). Then, the circles that best fit the pupil and iris are found using the Least Mean Square (LMS) algorithm~\cite{valenzuela2020towards}.

Finally, the iris is unwrapped using polar coordinates, as illustrated in Figure~\ref{fig:rubb}. We used the open-source method from Fang et al.~\cite{fang2020open} for this step. The result is the normalised iris image, also known as a rubber-sheet (Figure~\ref{fig:rubb}\textcolor{blue}{e}). All the rubber-sheets in Table~\ref{tab:dataset} were created with the same pipeline.

To reproduce this method, the full implementation can be accessed on GitHub\footnote{\url{https://github.com/Choapinus/DenseNet10}}. 

\subsection{Morph}

\subsubsection{Pairs selection}

The selection of iris pairs to create an iris morph is one of the most challenging processes. In morphing face images, state-of-the-art features are often extracted using a pre-trained network such as FaceNet, Arcface, or others. After that, the cosine similarity is used to look for pairs with similar scores ($1 - cos\ distance$) based on gender, age, and ethnicity, and the face morph image is created. However, these same conditions cannot be applied directly to iris images. The main concern is only separating the left and right eyes. Determining the best method to combine two irises for morph images is still an open challenge. 

For this work, we explored random selection and selection based on radius distance for pair selection.

\subsubsection{Generated morph images}


This work uses the already available databases ND-LG4000 and CASIA-IrisV4-Lamp, and while we do not focus on the database creation process (initial capture, eye detection, and cropping), and concentrate instead on the creation of iris morphed images, it is relevant to understand some previous stages in order to get a general-purpose view. The iris morphing method is based on face morphing algorithms, emphasising the following stages:

\begin{itemize}
    \item {\emph{Correspondence~\cite{sharma2021image}}}: Unlike the face, where features such as eyes, nose, mouth, and face contour can be readily used as landmarks, a different approach is needed in the case of the iris, where it is difficult to extract a fixed set of landmark points easily. We first obtain iris segmentation parameters based on the DenseNet10 CNN and the Least Mean Square (LMS) algorithm to establish the correspondence between two iris images. The iris centre, iris radius, pupil centre, and pupil radius are necessary to perform this stage. We estimate equally spaced landmarks on inner and outer iris boundaries using the segmentation parameters. The landmark points are 10 degrees apart with respect to the iris centre, resulting in 72 landmarks (36 on the inner iris boundary + 36 on the outer iris boundary). We select these 72 landmarks to minimise iris feature distortion during warping. A lower number of landmarks distorts the iris pattern during warping, and a higher number increases computational complexity. We also include four extreme corner points of an image (top left, top right, bottom left, and bottom right) in the landmarks set, creating a total of 76 landmark points. The corner points are required to align the iris regions of both images.
    \item \emph{Warping}: Given the landmark points, we compute the Delaunay triangulation using the convex hull method. We average the coordinates of the vertices of the corresponding triangles and derive the affine transformation matrix, $T = AX^{-1}$ as described in Equation~\ref{eq:matrix}.
    \begin{equation}\label{eq:matrix}
        \begin{bmatrix}
            t_1 & t_2 & t_3 \\
            t_4 & t_5 & t_6 \\
              0 &   0 &   1
        \end{bmatrix} =
        \begin{bmatrix}
            a_1 & b_1 & c_1 \\
            a_2 & b_2 & c_2 \\
              1 &   1 &   1
        \end{bmatrix}
        \begin{bmatrix}
            x_1 & y_1 & z_1 \\
            x_2 & y_2 & z_2 \\
              1 &   1 &   1
        \end{bmatrix}^{-1}
    \end{equation}
    
    Here, $A$ is the averaged triangle coordinates arranged column-wise, and $X$ is the set of coordinates from one of the corresponding triangles. Each transformation matrix is then used to map the pixels in the original triangle to the averaged triangle. Bilinear interpolation is used to populate missing pixel values in the averaged triangle.
    \item \emph{Blending}: Finally, we blend the pixels within the two warped triangles using linear blending at each location $(i, j)$ as $M(i, j) = \alpha Xw(i, j) + (1 - \alpha) Yw(i, j)$. 
    
    Here, $M$ is the morphed triangle, $Xw$ and $Yw$ are the two corresponding warped triangles, and $\alpha$ is the blending factor. The blending factor is set to $0.5$ to get an equal contribution from both images.
\end{itemize}

In the correspondence step, a set of correlated landmark points from both images are detected using the \emph{dlib} library with 68 landmark points~\cite{king2009dlib}. In the warping step, two images are non-linearly deformed to make them geometrically aligned with respect to the detected landmarks. 

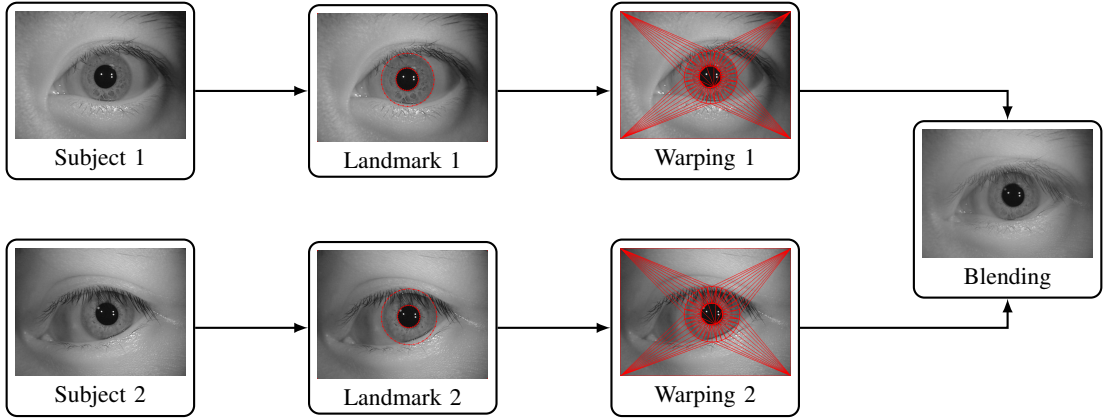
\begin{figure*}[!htb]
    \centering
    \begin{tikzpicture}[font=\small, thick]
        \node[draw, rounded corners, align=center] (block1) {\includegraphics[scale=0.1]{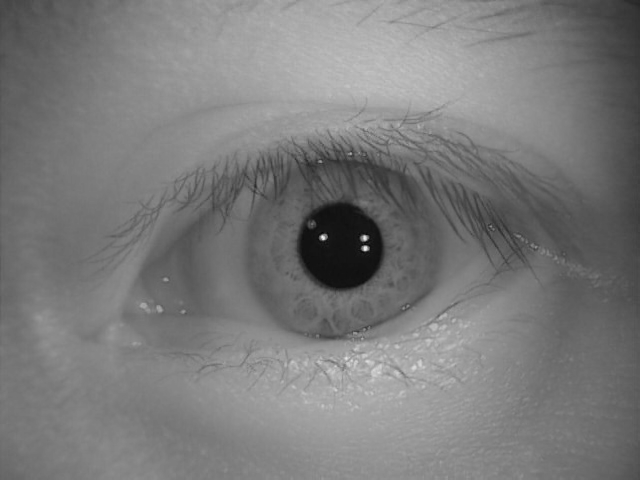}\\Blending};
        \node[draw, rounded corners, below left=-0.8cm and 1.5cm of block1, align=center] (block2) {\includegraphics[scale=0.1]{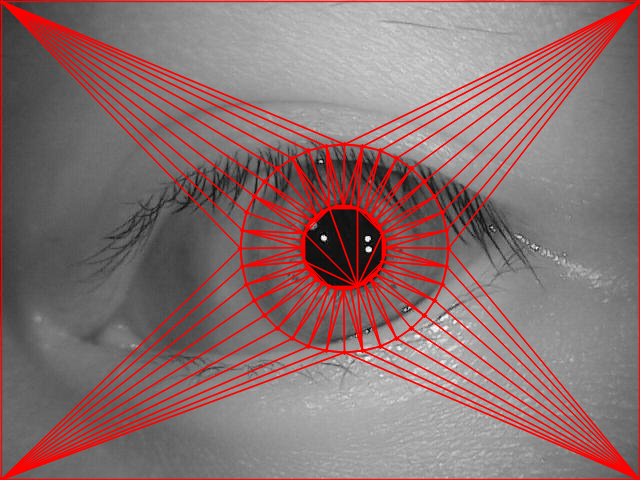}\\Warping 2};
        \node[draw, rounded corners, above left=-0.8cm and 1.5cm of block1, align=center] (block3) {\includegraphics[scale=0.1]{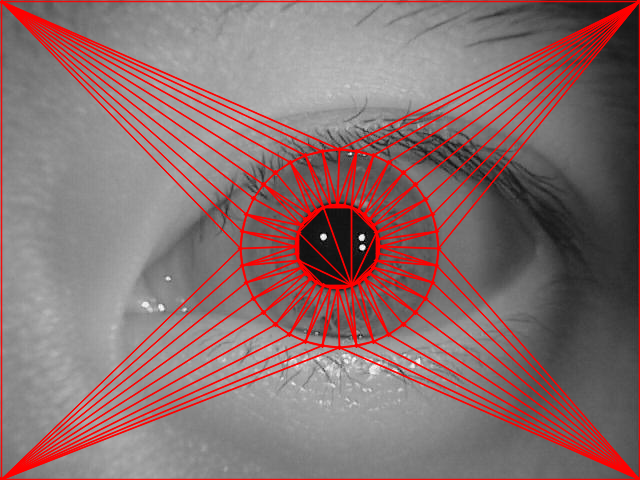}\\Warping 1};
        \node[draw, rounded corners, left=-0.8cm and 1.5cm of block2, align=center] (block4) {\includegraphics[scale=0.1]{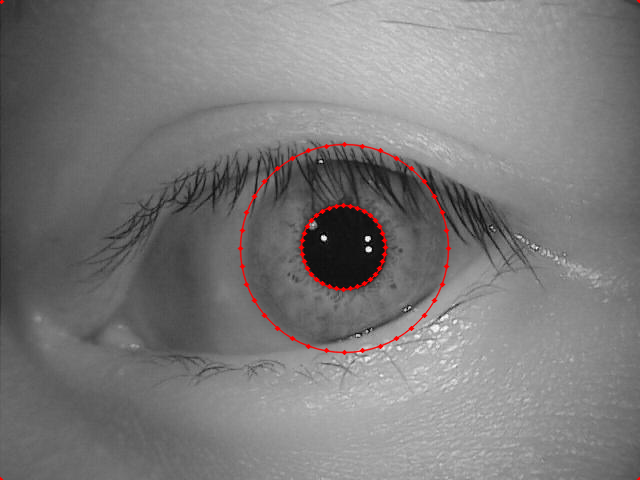}\\Landmark 2};
        \node[draw, rounded corners, left=-0.8cm and 1.5cm of block3, align=center] (block5) {\includegraphics[scale=0.1]{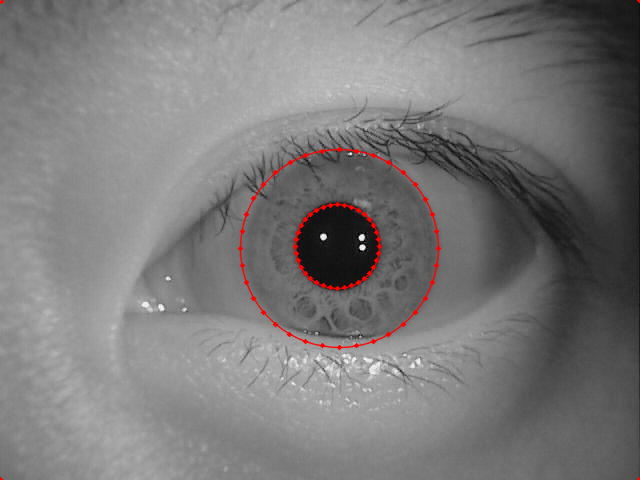}\\Landmark 1};
        \node[draw, rounded corners, left=-0.8cm and 1.5cm of block4, align=center] (block6) {\includegraphics[scale=0.1]{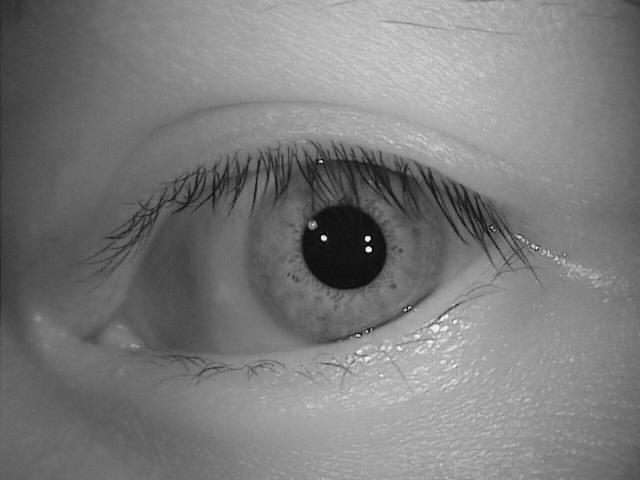}\\Subject 2};
        \node[draw, rounded corners, left=-0.8cm and 1.5cm of block5, align=center] (block7) {\includegraphics[scale=0.1]{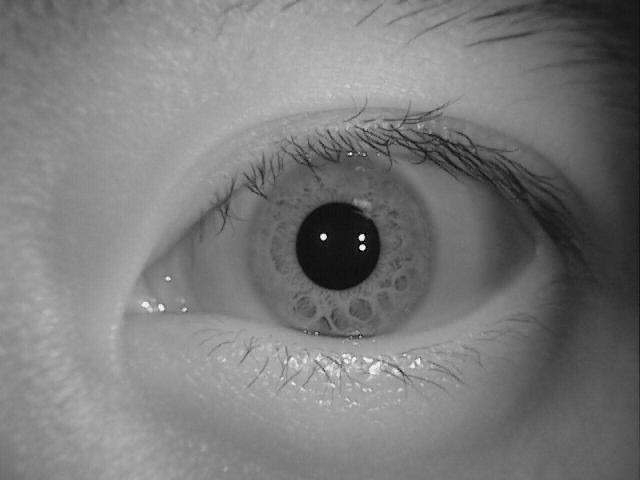}\\Subject 1};
        \draw[-latex] (block2) -| (block1);
        \draw[-latex] (block3) -| (block1);
        \draw[-latex] (block4) edge (block2);
        \draw[-latex] (block5) edge (block3);
        \draw[-latex] (block6) edge (block4);
        \draw[-latex] (block7) edge (block5);
    \end{tikzpicture}
    \caption{Iris morphing image generation process.}
    \label{fig:morph_process}
\end{figure*}

\begin{figure}[!htb]
    \centering
    \subfloat{\includegraphics[scale=0.12]{Figures/ND/bonafide/L02463d1890.png}}\hspace{0.8em}%
    \subfloat{\includegraphics[scale=0.12]{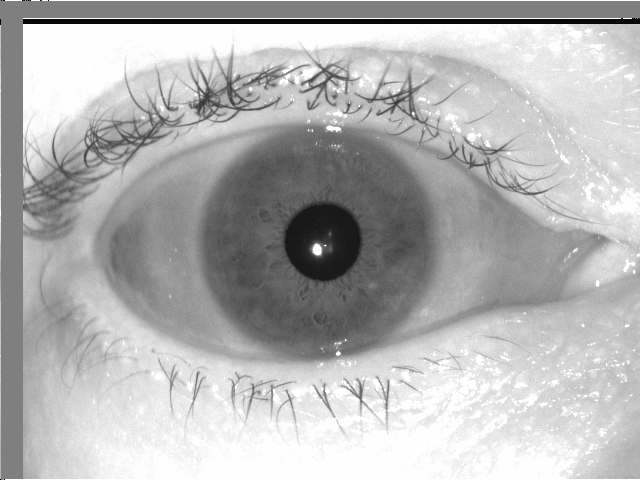}}\hspace{0.8em}%
    \subfloat{\includegraphics[scale=0.12]{Figures/ND/bonafide/L05691d121.png}}\\
    \subfloat{\includegraphics[scale=0.12]{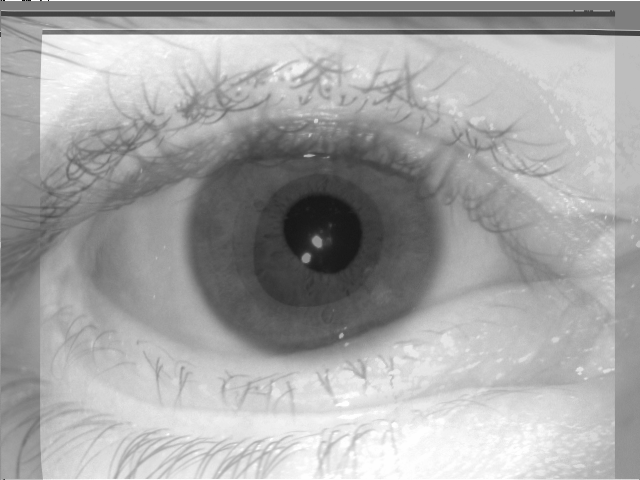}}\hspace{0.8em}%
    \subfloat{\includegraphics[scale=0.12]{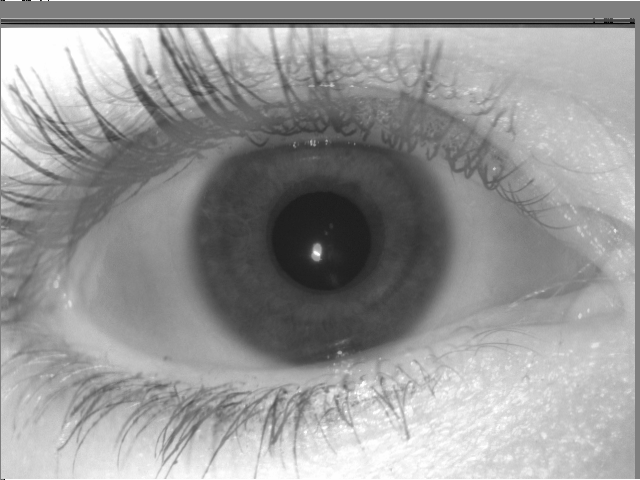}}\hspace{0.8em}%
    \subfloat{\includegraphics[scale=0.12]{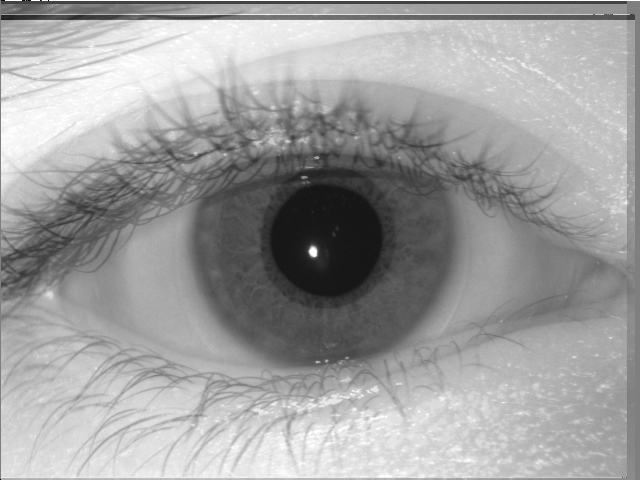}}\\
    \subfloat{\includegraphics[scale=0.12]{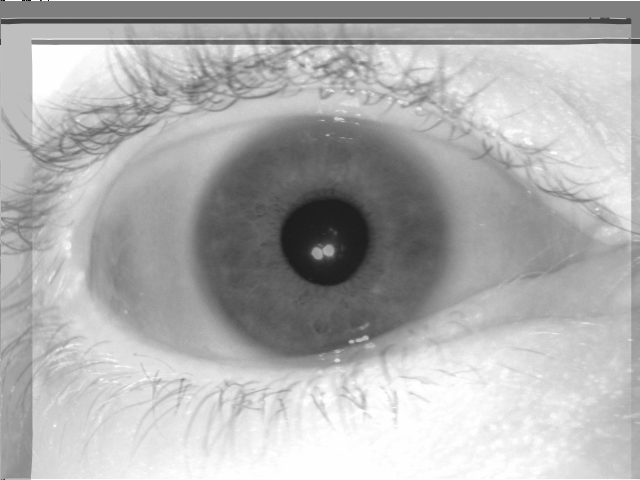}}\hspace{0.8em}%
    \subfloat{\includegraphics[scale=0.12]{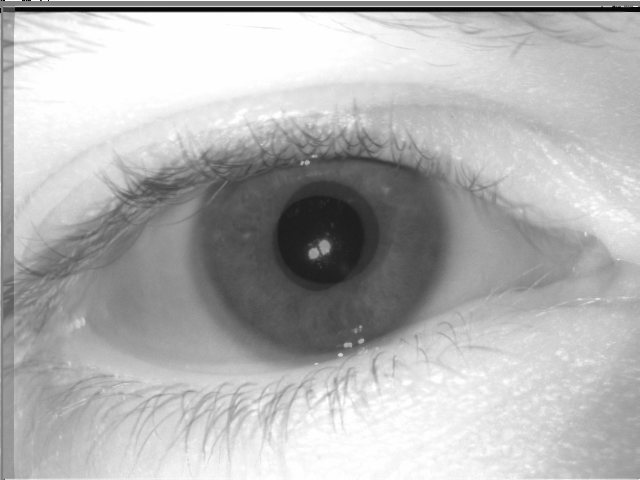}}\hspace{0.8em}%
    \subfloat{\includegraphics[scale=0.12]{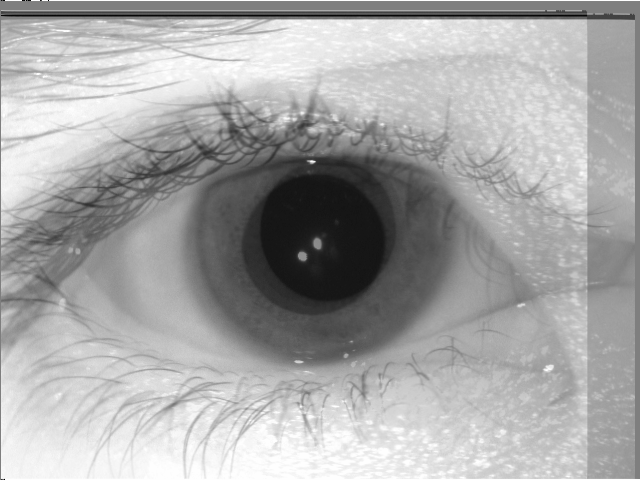}}
    \caption{Example of periocular morph images. Top: UND bona fide periocular images. Middle: Morph images using random selection. Bottom: Morph images using the radii pupil size method.}
    \label{fig:morph-example}
\end{figure}

Finally, the warped images are blended by linearly combining pixel values from both images at each location using a scalar value (blending factor). The scalar value controls the degree of contribution of each source image to the morphed image. Figure~\ref{fig:morph_process} shows the iris morph generation process. Figure~\ref{fig:morph-example} shows examples of bona fide and morph images.

Experimental evaluations are carried out using the Near-Infrared Range(NIR) LG-4000 sensor, and in the new periocular morph images. Figure~\ref{fig:morph-example} depicts sample images of the used dataset. From these, periocular images can be generated, normalising and encoding images for traditional approaches. Then, the iris morph at the image level allows us multiple evaluations.

Based on the observation that different eyes tend to be more similar when originating from the same eye position, we process only images of all 487 left and 487 right eyes. A total number of 66,114 mated and 1,079,445 non-mated iris codes are created from pairs of the first image of each subject.

In an attack attempt, two different instances of iris-codes contributing to a morphed iris-code are compared against it. An attack attempt is considered successful if the larger of the two obtained Hamming distance scores is below the decision threshold, i.e., $\max(HD(CodeM, CodeA'), HD(CodeM, CodeB')) \leq \delta$. We consider decision thresholds at FMR of 0.01\% and 0.001\%, which is frequently reported in iris recognition research. Furthermore, we consider a Hamming distance of 0.32 as the decision criterion recommended in~\cite{rathgeb2017feasibility}. Depending on the number of remaining iris images, up to five comparisons are performed against each morphed iris code, resulting in 16,668 attack attempts. Results are reported in Section~\ref{sec:experiments}.

\section{Experiments and Results}
\label{sec:experiments} 

The experiments described in this section aim to compare the similarity between bona fide and periocular morph images, as well as evaluate the impact of using those morphed images in a state-of-the-art iris recognition system based on DenseNet201~\cite{zambrano2022iris}. 

\subsection{Iris recognition}

For both backbones used in the Siamese network, the input images are resized to $224\times 224$ before being processed by the network. The size of the embeddings is $1\times 2,048$ for ResNet50 and $1\times 1,792$ for MobileNetV2.

Both backbone networks were fine-tuned from the available ImageNet weights. For the ResNet50 backbone, the following hyperparameters were used: unfreezing from layer \enquote{conv5\_block2\_out} after a grid search process, a batch size of $1,024$, a learning rate of $1\times 10^{-5}$, $400$ epochs, and Adam optimisation. 

For the MobileNetV2 backbone, the best results were obtained using the following hyperparameters: alpha parameter multiplication of $1.4$, unfreezing from layer \enquote{block\_16\_expand} after a grid search process, batch size of $512$, a learning rate of $1\times 10^{-5}$, $300$ epochs and Adam optimisation. The best performing SiamIris model will be made publicly available on GitHub\footnote{\url{https://github.com/dpbenalcazar/SiamIris-v1}} upon acceptance. 

\begin{figure*}[!htb]
    \centering
    \subfloat{\includegraphics[scale=0.40]{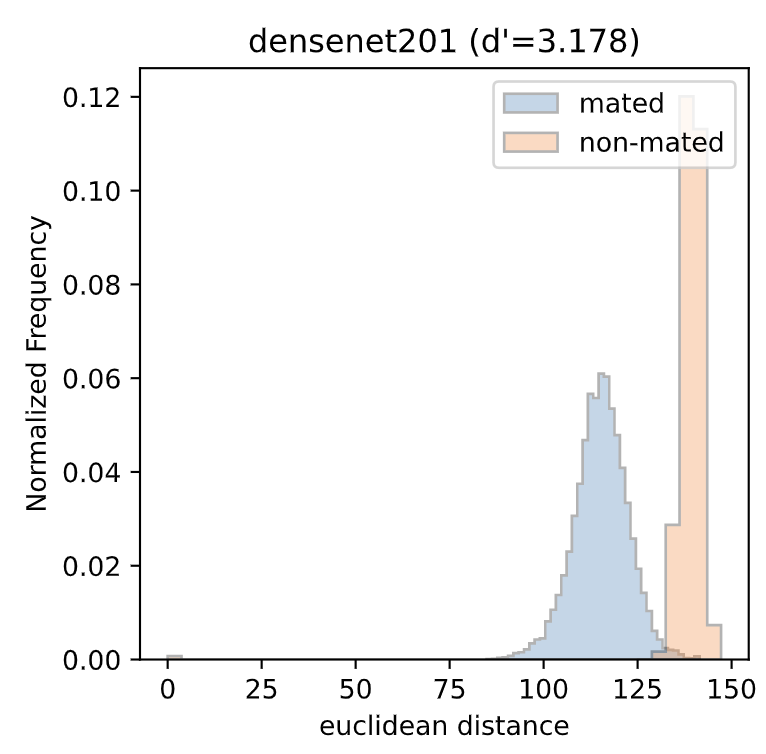}}\hfil%
    \subfloat{\includegraphics[scale=0.40]{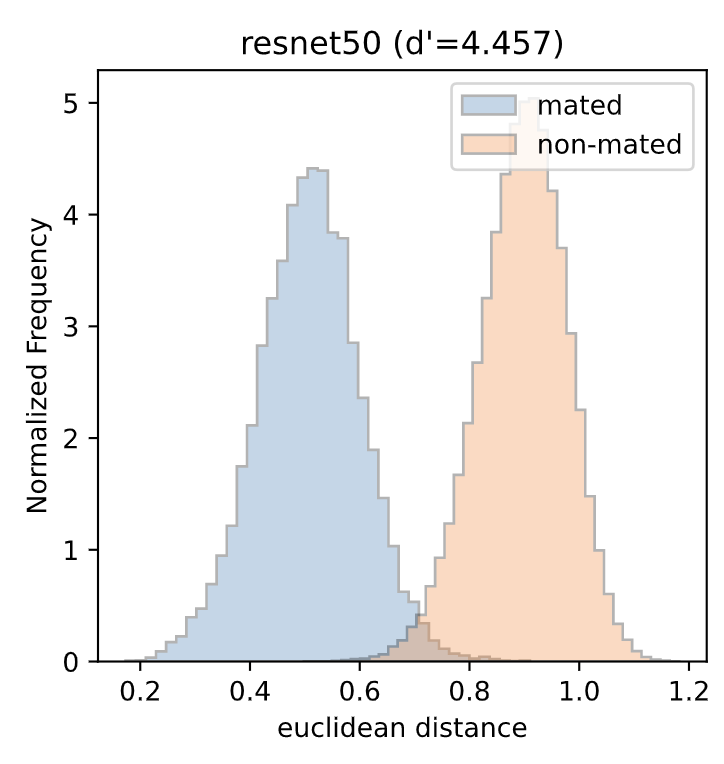}}\hfil%
    \subfloat{\includegraphics[scale=0.40]{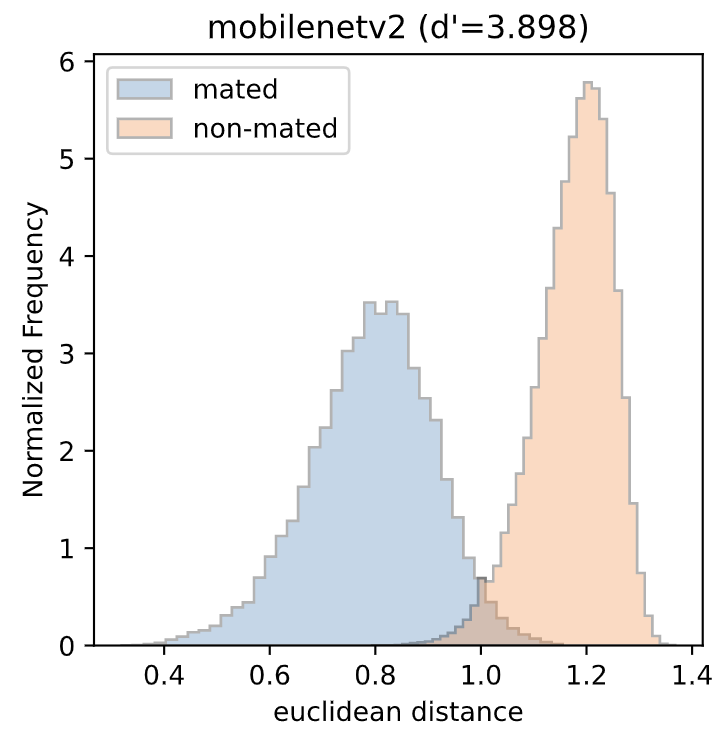}}
    \caption{Histogram $d'$ value comparison between mated and non-mated distributions for DenseNet201 (state of the art), SiamIris-R50-periocular (ResNet50 backbone), and SiamIris-MN-periocular (MobileNetV2 backbone). The Y-axis uses a normalised frequency in order to normalise all the distributions to the range between 0 and 1.}
    \label{fig:d_plot}
\end{figure*}

Figure~\ref{fig:d_plot} shows the results of the $d'$ metrics comparison for DenseNet201, SiamIris R50-periocular and SiamIris MN-periocular. It is essential to highlight that a higher $d'$ value means a better separation distance between mated and non-mated. The Y-Axis in the plot uses a normalised frequency because the distributions were normalised to the range between 0 and 1. Both Siamese models obtained better $d'$ than DenseNet201.

Table~\ref{tab:results} shows the comparison results among our proposed methods (SiamIris-R50-periocular and SiamIris-MN-periocular), and the best results in the state-of-the-art based on DenseNet201. Our proposal based on periocular images performs better than the models based on rubber-sheet images and DenseNet201, especially at FNMR\tu{10} and FNMR\tu{20} operating points.

\begin{table}[!htb]\def\tabularxcolumn#1{m{#1}}
    \centering
    \caption{Results on the ND-LG4000-LR Dataset for Iris Recognition DL methods using periocular images.}
    \label{tab:results}
    \setlength{\tabcolsep}{4pt} 
    \begin{tabularx}{\linewidth}{lrYYY}
        \toprule
        \textbf{Method} & \textbf{d'[$\uparrow$]} & 
        \textbf{EER (\%)} & \textbf{FMR \tu{@FNMR=10} ([$\downarrow$] \%)} & \textbf{FNMR \tu{@FNMR=20} ([$\downarrow$] \%)} \\
        \midrule
        DenseNet201~\cite{zambrano2022iris} &          3.17 &          0.98 &          0.32 &          0.53 \\
        \textbf{SiamIris R50-peri.}         & \textbf{4.45} & \textbf{1.57} & \textbf{0.26} & \textbf{0.51} \\
        SiamIris R50-rubb                   &          4.13 &          2.38 &          0.51 &          1.12 \\ 
        SiamIris MN2-peri.                  &          3.89 &          2.59 &          0.42 &          1.01 \\
        SiamIris MN2-rubb                   &          3.80 &          2.94 &          0.68 &          1.63 \\
        \bottomrule
    \end{tabularx}
\end{table}

Table~\ref{tab:po_nd_non_mated} reports the result of SiamIris for periocular morph images for mated and non-mated distributions. The FNMR is reported at three different security levels (thresholds): 10\%, 5\%, and 1\%, respectively.

\begin{table}[!htb]\def\tabularxcolumn#1{m{#1}}
    \centering
    \caption{SiamIris: periocular morph images. Top: NotreDame-LG4000-LR, bottom: CASIA-IrisV4}
    \label{tab:po_nd_non_mated}
    \setlength{\tabcolsep}{4pt}

    \begin{tabularx}{\linewidth}{lrYYY}
        \toprule
        Metrics & \textbf{EER} & \textbf{FMR \tu{@FNMR=10\%}} & 
        \textbf{FMR \tu{@FNMR=5\%}} & \textbf{FMR \tu{@FNMR=1\%}} \\
        \midrule
        \emph{NotreDame-LG4000-LR} & ($d'$ 4.46) & & & \\
        \midrule
        \textbf{\emph{Threshold ($\tau$)}} &   \emph{0.70} &   \emph{0.78} &   \emph{0.75} &   \emph{0.69} \\
        \textbf{FNMR (\%)}                 &          1.57 &          0.26 &          0.51 &          2.47 \\
        \midrule
        \emph{CASIA-IrisV4} & ($d'$ 3.33) & & & \\
        \midrule
        \textbf{\emph{Threshold ($\tau$)}} &   \emph{0.52} &   \emph{0.55} &   \emph{0.52} &   \emph{0.47} \\
        \textbf{FNMR (\%)}                 &          5.24 &          3.39 &          5.39 &         11.31 \\
        \bottomrule
    \end{tabularx}
\end{table}

Table~\ref{tab:rb_nd_non_mated} reports the result of SiamIris for both mated and non-mated images using the rubber-sheet as input. The FNMR is also reported at three different thresholds: 10\%, 5\%, and 1\%, respectively.

\begin{table}[!htb]\def\tabularxcolumn#1{m{#1}}
    \centering
    \caption{SiamIris: rubber-sheet images using radius pair selection. Top: NotreDame-LG4000-LR, bottom: CASIA-IRISV4}
    \label{tab:rb_nd_non_mated}
    \setlength{\tabcolsep}{4pt}
    \begin{tabularx}{\linewidth}{lrYYY}
        \toprule
        Metrics & \textbf{EER} & \textbf{FMR \tu{@FNMR=10\%}} & \textbf{FMR \tu{@FNMR=5\%}} & \textbf{FMR \tu{@FNMR=1\%}} \\
        \midrule
        \emph{NotreDame-LG4000-LR} & ($d'$ 6.32) & & & \\
        \midrule
        \textbf{\emph{Threshold ($\tau$)}} &   \emph{0.33} &   \emph{0.38} &   \emph{0.37} &   \emph{0.35} \\
        \textbf{FNMR (\%)}                 &          0.22 &          0.15 &          0.16 &          0.19 \\
        \midrule
        \emph{CASIA-IrisV4} & ($d'$ 2.79) & & & \\
        \midrule
        \textbf{\emph{Threshold ($\tau$)}} &   \emph{0.35} &   \emph{0.35} &   \emph{0.34} &   \emph{0.32} \\
        \textbf{FNMR (\%)}                 &          8.52 &          8.09 &          9.92 &         14.09 \\
        \bottomrule
    \end{tabularx}
\end{table}

\begin{figure*}[!htb]
    \centering
    \subfloat[\emph{NotreDame-LG4000-LR} radius selection histogram]{\includegraphics[width=0.24\linewidth]{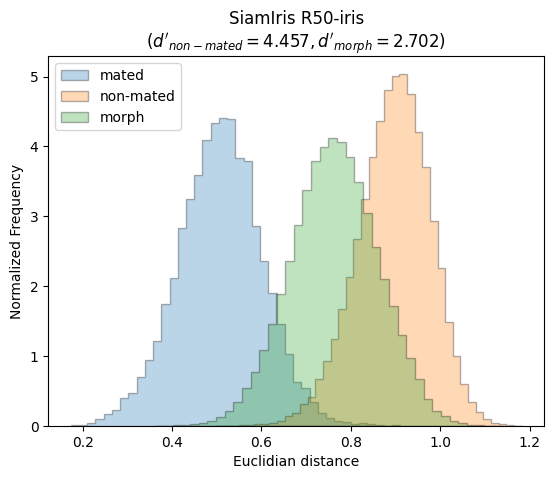}}\hfil%
    \subfloat[\emph{NotreDame-LG4000-LR} radius selection DET Curve]{\includegraphics[width=0.24\linewidth]{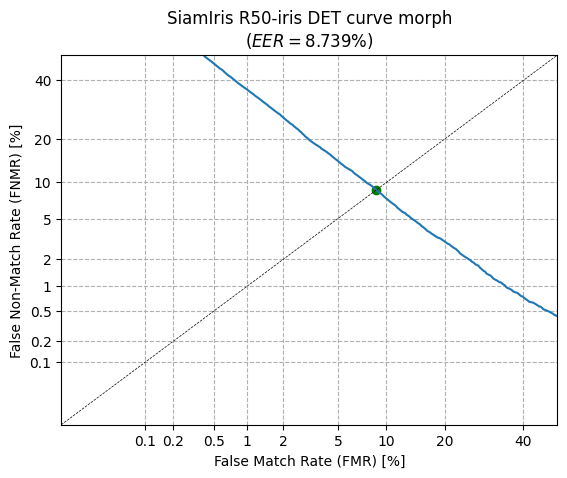}}\hfil%
    \subfloat[\emph{NotreDame-LG4000-LR} random selection histogram]{\includegraphics[width=0.24\linewidth]{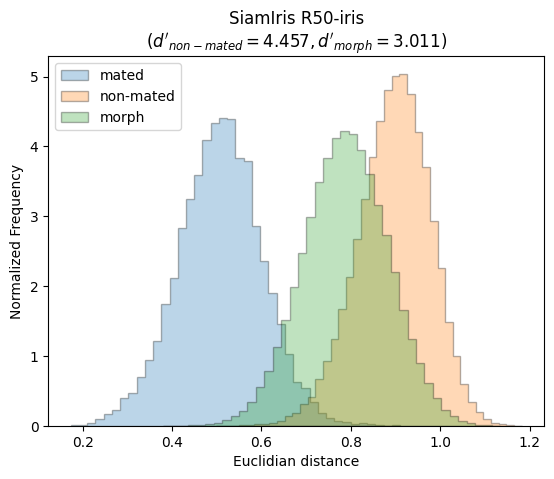}}\hfil%
    \subfloat[\emph{NotreDame-LG4000-LR} random selection DET Curve]{\includegraphics[width=0.24\linewidth]{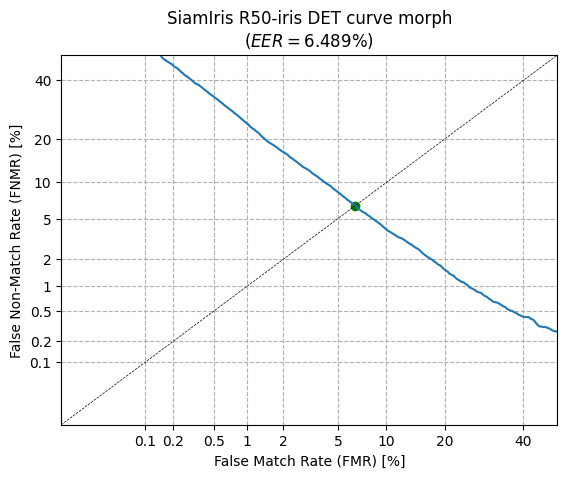}}\\
    \subfloat[\emph{CASIA-IrisV4} radius selection histogram]{\includegraphics[width=0.24\linewidth]{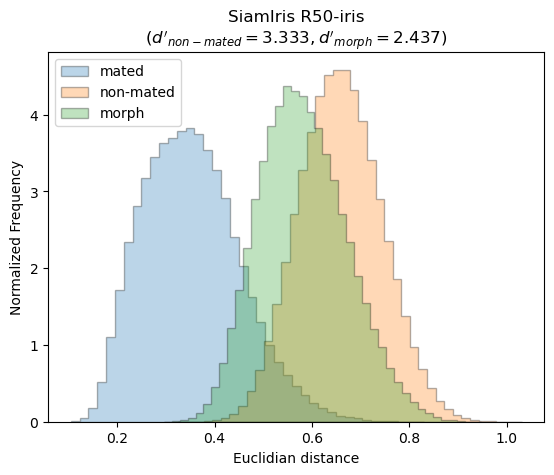}}\hfil%
    \subfloat[\emph{CASIA-IrisV4} radius selection DET Curve]{\includegraphics[width=0.24\linewidth]{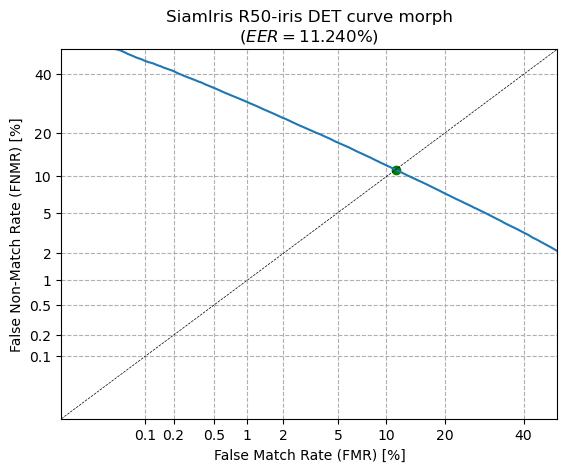}}\hfil%
    \subfloat[\emph{CASIA-IrisV4} random selection histogram]{\includegraphics[width=0.24\linewidth]{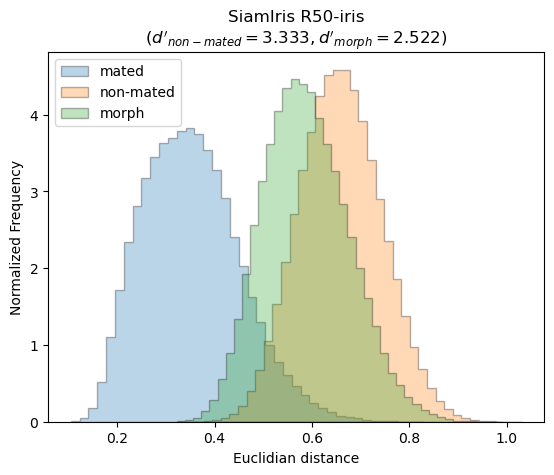}}\hfil%
    \subfloat[\emph{CASIA-IrisV4} random selection DET Curve]{\includegraphics[width=0.24\linewidth]{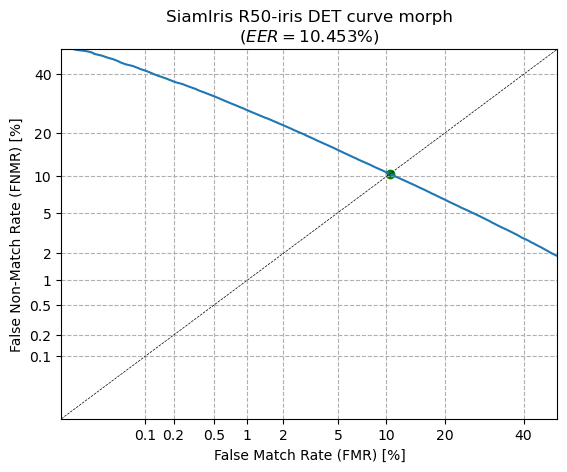}}
    \caption{SiamIris using \textbf{periocular} images. All histograms show the $d'$ value for mated, no-mated and morph images. All DET curves show results for mated vs morph. (a), (b), (c), (d) show results for the \emph{NotreDame-LG4000-LR} database. (a) shows the Histogram using radius pair selection, (b) shows the DET curve using radius pair selection, (c) shows the Histogram using random pair selection, and (d) shows the DET curve using random pair selection. (e), (f), (g), (h) show results for the \emph{CASIA-IrisV4} database. (e) shows the Histogram using radius pair selection, (f) shows the DET curve using radius pair selection, (g) shows the Histogram using random pair selection, and (h) shows the DET curve using random pair selection.}
    \label{fig:po_cs_morph_radius}
\end{figure*}

Figure~\ref{fig:po_cs_morph_radius} shows the impact on periocular morph images in the proposed iris recognition system, with the best performance based on SiamIris for both datasets ND-LG4000-LR and CASIA-IrisV4. The histogram shows the feasibility of being attacked with success when iris morph images were created using periocular images.

\begin{figure*}[!htb]
    \centering
    \subfloat[\emph{NotreDame-LG4000-LR} radius selection histogram]{\includegraphics[width=0.24\linewidth]{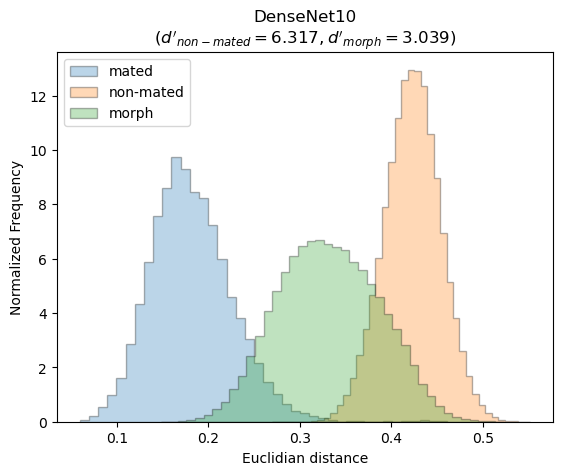}}\hfil%
    \subfloat[\emph{NotreDame-LG4000-LR} radius selection DET Curve]{\includegraphics[width=0.24\linewidth]{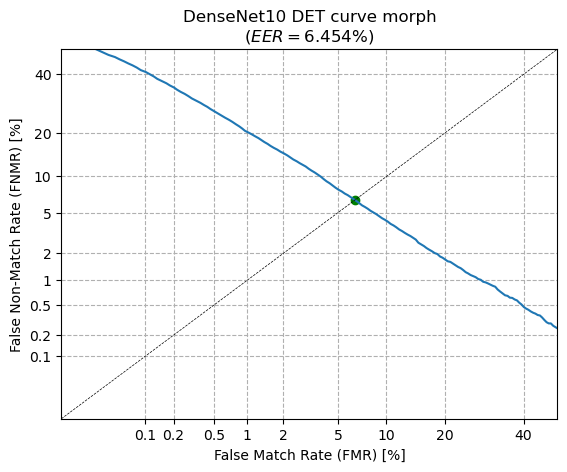}}\hfil%
    \subfloat[\emph{NotreDame-LG4000-LR} random selection histogram]{\includegraphics[width=0.24\linewidth]{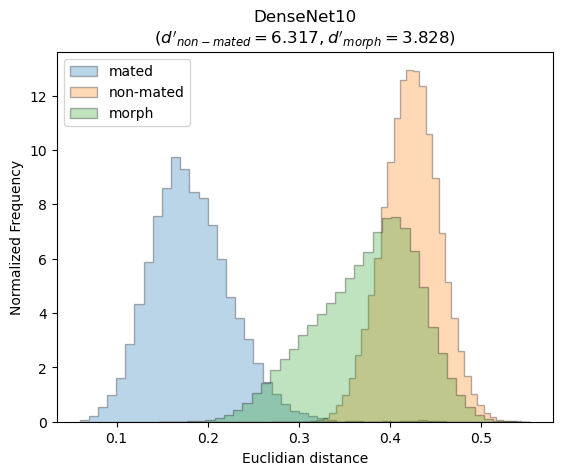}}\hfil%
    \subfloat[\emph{NotreDame-LG4000-LR} random selection DET Curve]{\includegraphics[width=0.24\linewidth]{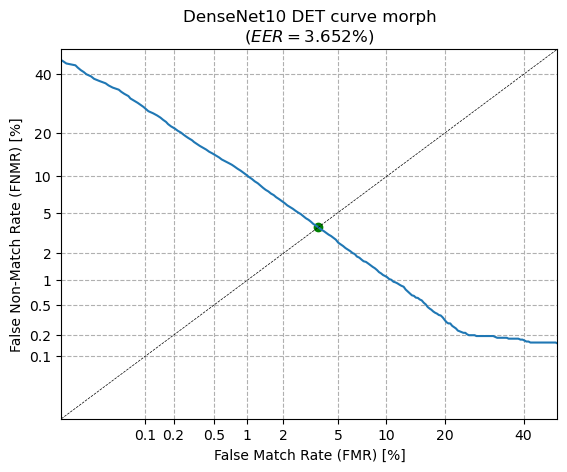}}\\
    \subfloat[\emph{CASIA-IrisV4} radius selection histogram]{\includegraphics[width=0.24\linewidth]{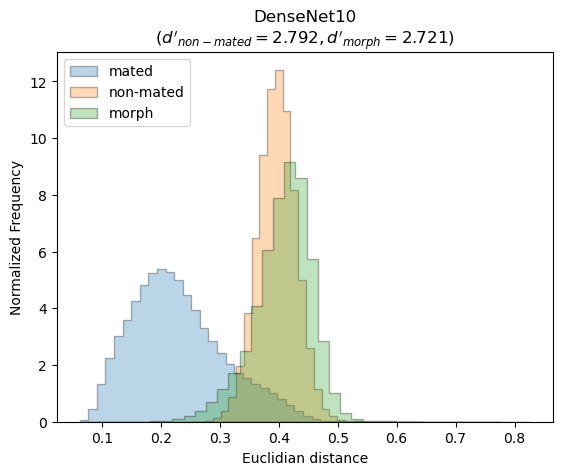}}\hfil%
    \subfloat[\emph{CASIA-IrisV4} radius selection DET Curve]{\includegraphics[width=0.24\linewidth]{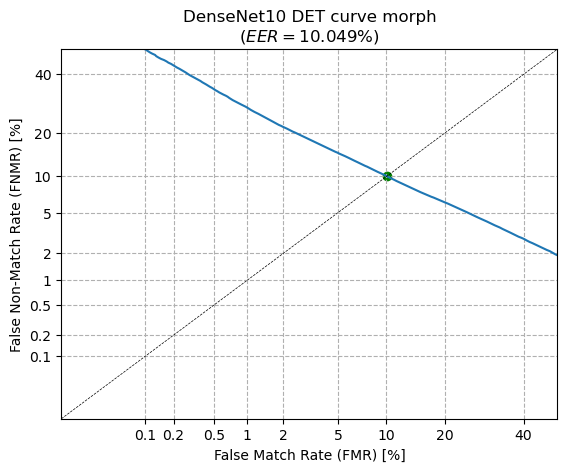}}\hfil%
    \subfloat[\emph{CASIA-IrisV4} random selection histogram]{\includegraphics[width=0.24\linewidth]{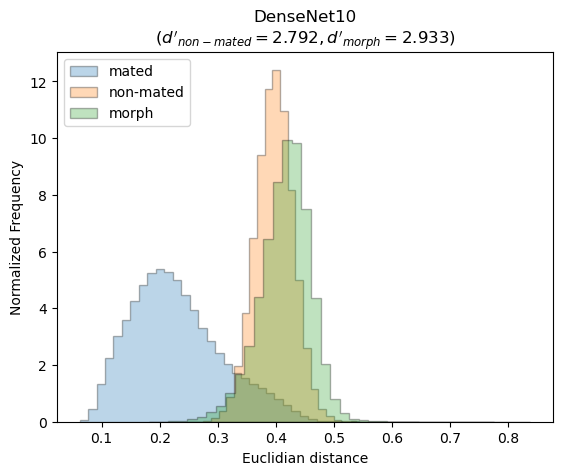}}\hfil%
    \subfloat[\emph{CASIA-IrisV4} random selection DET Curve]{\includegraphics[width=0.24\linewidth]{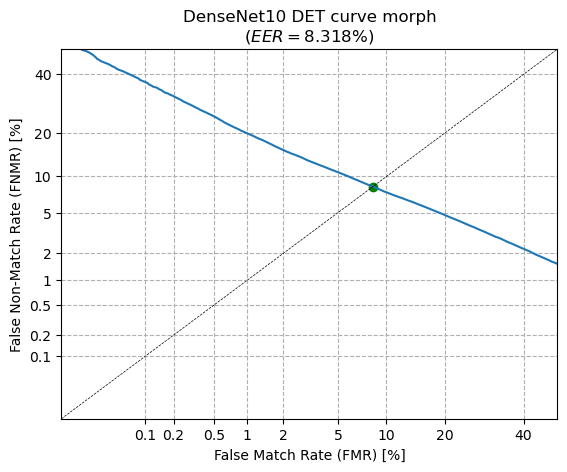}}
    \caption{SiamIris using \textbf{rubber-sheets} obtained from perioucular images. All histograms show the $d'$ value for mated, no-mated and morph images. All DET curves show results for mated vs morph. (a), (b), (c), (d) show results for the \emph{NotreDame-LG4000-LR} database. (a) shows the Histogram using radius pair selection, (b) shows the DET curve using radius pair selection, (c) shows the Histogram using random pair selection, and (d) shows the DET curve using random pair selection. (e), (f), (g), (h) show results for the \emph{CASIA-IrisV4} database. (e) shows the Histogram using radius pair selection, (f) shows the DET curve using radius pair selection, (g) shows the Histogram using random pair selection, and (h) shows the DET curve using random pair selection.}
    \label{fig:rb_nd_morph_radius}
\end{figure*}

Figure~\ref{fig:rb_nd_morph_radius} shows the effect on periocular morph images in the proposed iris recognition system, with the best performance based on SiamIris for both datasets ND-LG4000-LR and CASIA-IrisV4. The histogram shows the feasibility of being attacked with success when iris morph images were created using the rubber-sheet obtained from morph images.

\subsection{Vulnerability Analysis}

We evaluated the iris periocular images to show how the morph images impact the iris recognition system. It is essential to highlight that creating morph periocular images allows us to obtain the three traditional representations of the iris biometric sample. Then, other researchers could make the iris and texture images based only on the morphed periocular iris images.

As already mentioned, the MAP can be used to assess the vulnerability of the iris recognition system. 
The relative metric RMMR can integrate with the MAP. A significant downside of RMMR is that when the bounds of False Rejection Rate (FRR) tends to 100\%, the RMMR is greater than 100\%, which is not realistic. However, an IRS that provides an FRR equal to 100\% could not be considered to be a strong system either.


Table~\ref{tab:po_nd_morph_radius} shows the vulnerability assessment of iris recognition techniques for periocular iris morph attacks for the ND-LG4000 database. We report metrics at different thresholds, corresponding to FMR at 10\%, 5\%, and 1\% values of FNMR, respectively. These results represent radius pupil selection criteria.

\begin{table}[!htb]\def\tabularxcolumn#1{m{#1}}
    \centering
    \caption{Vulnerability analysis: Notre-Dame-LG4000-LR dataset; periocular images; radius pair selection.}
    \label{tab:po_nd_morph_radius}
    \begin{tabularx}{\linewidth}{lYYYY}
        \toprule
        $d'$ 2.7020 &&&& \\
        \midrule
        Metrics & \textbf{EER} & \textbf{FMR \tu{@FNMR=10\%}} & \textbf{FMR \tu{@FNMR=5\%}} & \textbf{FMR \tu{@FNMR=1\%}} \\
        \midrule
        \textbf{\emph{Threshold ($\tau$)}} & \emph{0.6319} & \emph{0.6395} & \emph{0.6041} & \emph{0.5397} \\
        \textbf{FNMR (\%)}                 &        8.7391 &        7.4623       & 14.3570       & 36.4864 \\
        \textbf{MinMax-RMMR}               &        0.9747 &        1.0261       &  1.1251       &  1.3623 \\
        \textbf{ProdAvg-RMMR}              &        0.8489 &        0.8842       &  1.0472       &  1.3452 \\
        \bottomrule
    \end{tabularx}
\end{table}

Table~\ref{tab:po_nd_morph_random} reports the vulnerability assessment of iris recognition using random selection criteria for periocular iris images. 

\begin{table}[!htb]\def\tabularxcolumn#1{m{#1}}
    \centering
    \caption{Vulnerability analysis: Notre-Dame-LG4000-LR dataset; periocular images; random pair selection.}
    \label{tab:po_nd_morph_random}
    \begin{tabularx}{\linewidth}{lYYYY}
        \toprule
        $d'$ 3.0111 &&&& \\
        \midrule
        Metrics & \textbf{EER} & \textbf{FMR \tu{@FNMR=10\%}} & \textbf{FMR \tu{@FNMR=5\%}} & \textbf{FMR \tu{@FNMR=1\%}} \\
        \midrule
        \textbf{\emph{Threshold ($\tau$)}} & \emph{0.6461} & \emph{0.6682} & \emph{0.6339} & \emph{0.5703} \\
        \textbf{FNMR (\%)}                 &        6.4886 &        3.9834 &        8.3609 &       24.5362 \\
        \textbf{MinMax-RMMR}               &        0.9905 &        0.9966 &        1.0644 &        1.2415 \\
        \textbf{ProdAvg-RMMR}              &        0.8894 &        0.8489 &        0.9855 &        1.2249 \\
        \bottomrule
    \end{tabularx}
\end{table}

Table~\ref{tab:po_cs_morph_radius}, again, shows the vulnerability assessment of iris recognition techniques for periocular iris morph attacks, but this time for the CASIA-IrisV4 database. We report metrics at different thresholds, corresponding to FMR at 10\%, 5\%, and 1\% values of FNMR, respectively. These results represent radius pupil selection criteria. 

\begin{table}[]\def\tabularxcolumn#1{m{#1}}
    \centering
    \caption{Vulnerability analysis: CASIA-IrisV4; periocular images; radius pair selection.}
    \label{tab:po_cs_morph_radius}
    \begin{tabularx}{\linewidth}{lYYYY}
        \toprule
        $d'$ 2.4365 &&&& \\
        \midrule
        Metrics & \textbf{EER} & \textbf{FMR \tu{@FNMR=10\%}} & \textbf{FMR \tu{@FNMR=5\%}} & \textbf{FMR \tu{@FNMR=1\%}} \\
        \midrule
        \textbf{\emph{Threshold ($\tau$)}} & \emph{0.4752} & \emph{0.4700} & \emph{0.4444} & \emph{0.4011} \\
        \textbf{FNMR (\%)}                 &       11.2401 &       12.1001 &       17.3434 &       29.6747 \\
        \textbf{MinMax-RMMR}               &        1.0384 &        1.1104 &        1.1711 &        1.2967 \\
        \textbf{ProdAvg-RMMR}              &        0.8525 &        0.9411 &        1.0795 &        1.2772 \\
        \bottomrule
    \end{tabularx}
\end{table}

Table~\ref{tab:po_cs_morph_random} reports the vulnerability assessment of iris recognition using random selection criteria for periocular morph images. It is essential to highlight again that for all Tables, the $d'$ factor reported is the distance related between the mated and morph distribution. A higher $d'$ factor means that morph images are easier to detect, and a low $d'$ factor implies that the two distributions are so close that it is hard to detect.

\begin{table}[]\def\tabularxcolumn#1{m{#1}}
    \centering
    \caption{Vulnerability analysis: CASIA-IrisV4; periocular images; random pair selection.}
    \label{tab:po_cs_morph_random}
    \begin{tabularx}{\linewidth}{lYYYY}
        \toprule
        \textbf{$d'$} & 2.5222 \\
        \midrule
        Metrics & \textbf{EER} & \textbf{FMR \tu{@FNMR=10\%}} & \textbf{FMR \tu{@FNMR=5\%}} & \textbf{FMR \tu{@FNMR=1\%}} \\
        \midrule
        \textbf{\emph{Threshold ($\tau$)}} & \emph{0.4800} & \emph{0.4783} & \emph{0.4527} & \emph{0.4095} \\
        \textbf{FNMR (\%)}                 & 10.4526       &       10.7427 &       15.5033 &       26.9835 \\
        \textbf{MinMax-RMMR}               & 1.0408        &        1.0971 &        1.1536 &        1.2698 \\
        \textbf{ProdAvg-RMMR}              & 0.8656        &        0.9282 &        1.0615 &        1.2502 \\
        \bottomrule
    \end{tabularx}
\end{table}


\subsection{Morphing Attack Detection}

An explicit Random Forest (RF) machine learning classifier was implemented to analyse the potential of the new morph images generated based on random and radius pair selection. The S-MAD approach was implemented to explore MAD capabilities.

The intensity values of greyscale pixels, and Discrete Cosine Transform (DCT) were explored based on a Discrete Fourier Transform for all the images based on results obtained in~\cite{tapia2023face} for face morphing. The DCT decomposes a discrete time-domain signal into its frequency components. 

Only the magnitude (real) and not the phase (complex) are used for training purposes. The magnitude image is then transformed from a linear scale to a logarithmic scale to compress the range of values. Furthermore, the quadrants of the matrix are shifted so that zero-value frequencies are placed at the centre of the image.

The dataset was divided into 70\% and 30\%, respectively, for training and testing. The set contains 2,400 bona fide images, 4,000 random selection morphs, and 4,000 radius selection morph images. 

First, the classifier was trained using bona fide and random selection pairs for morphing, and radius selection pairs were used as a test set. Afterwards, the classifier was trained using bona fide and radius selection pairs, and random selection morphed pairs were used as the test set.

\begin{figure*}[!htb]
    \centering
    \subfloat{\includegraphics[scale=0.15]{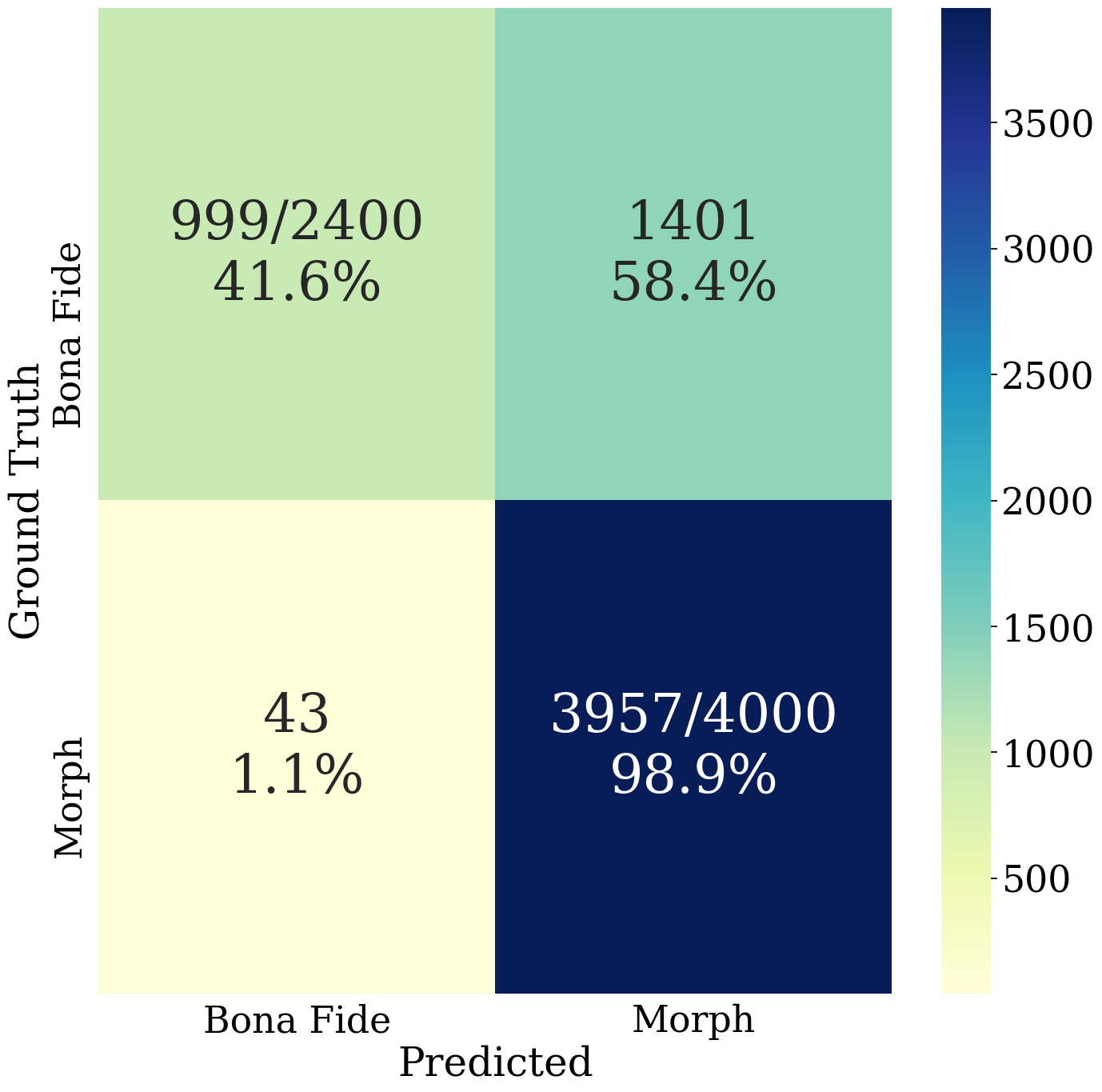}}\hfil%
    \subfloat{\includegraphics[scale=0.15]{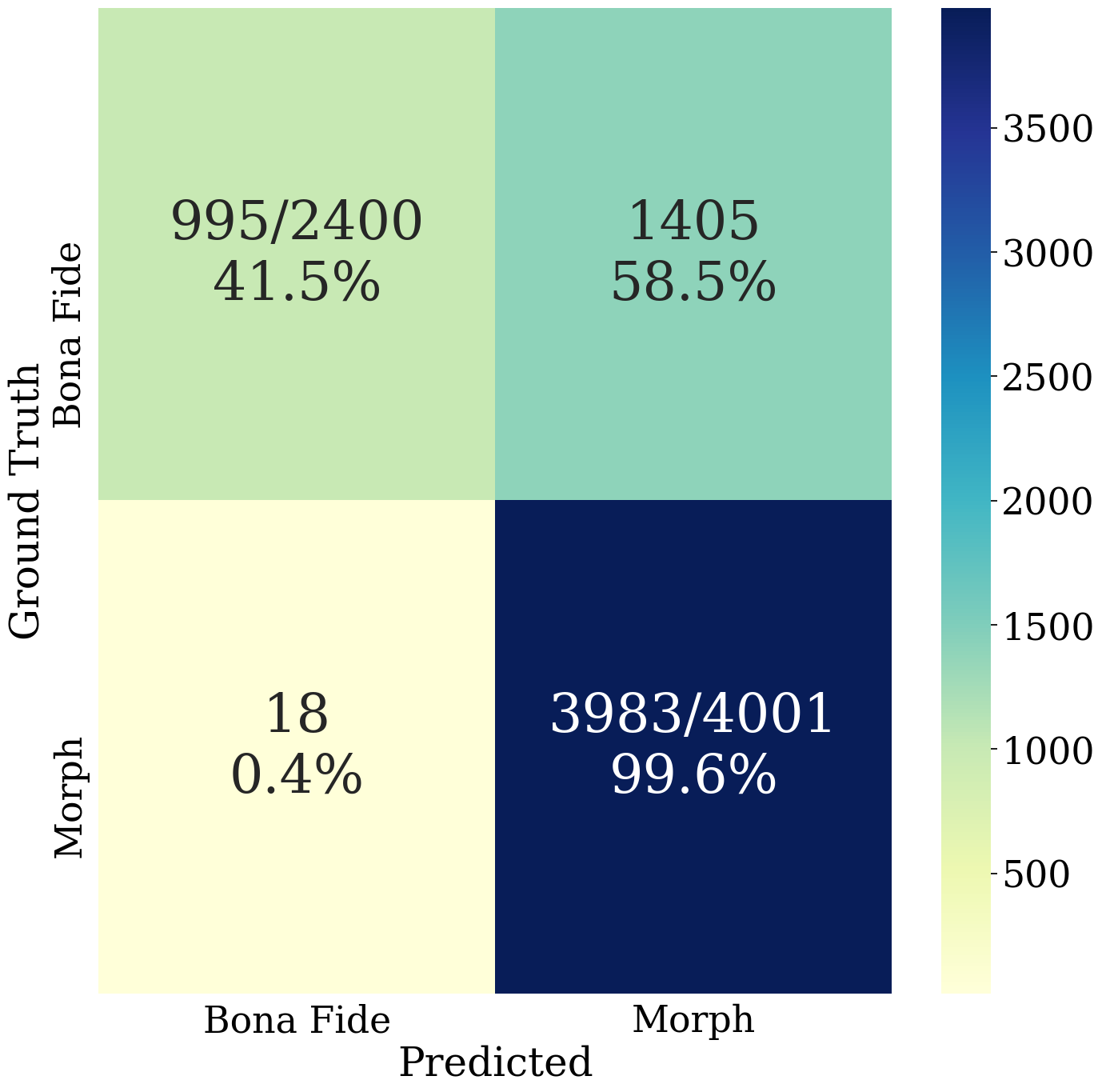}}\hfil%
    \subfloat{\includegraphics[scale=0.24]{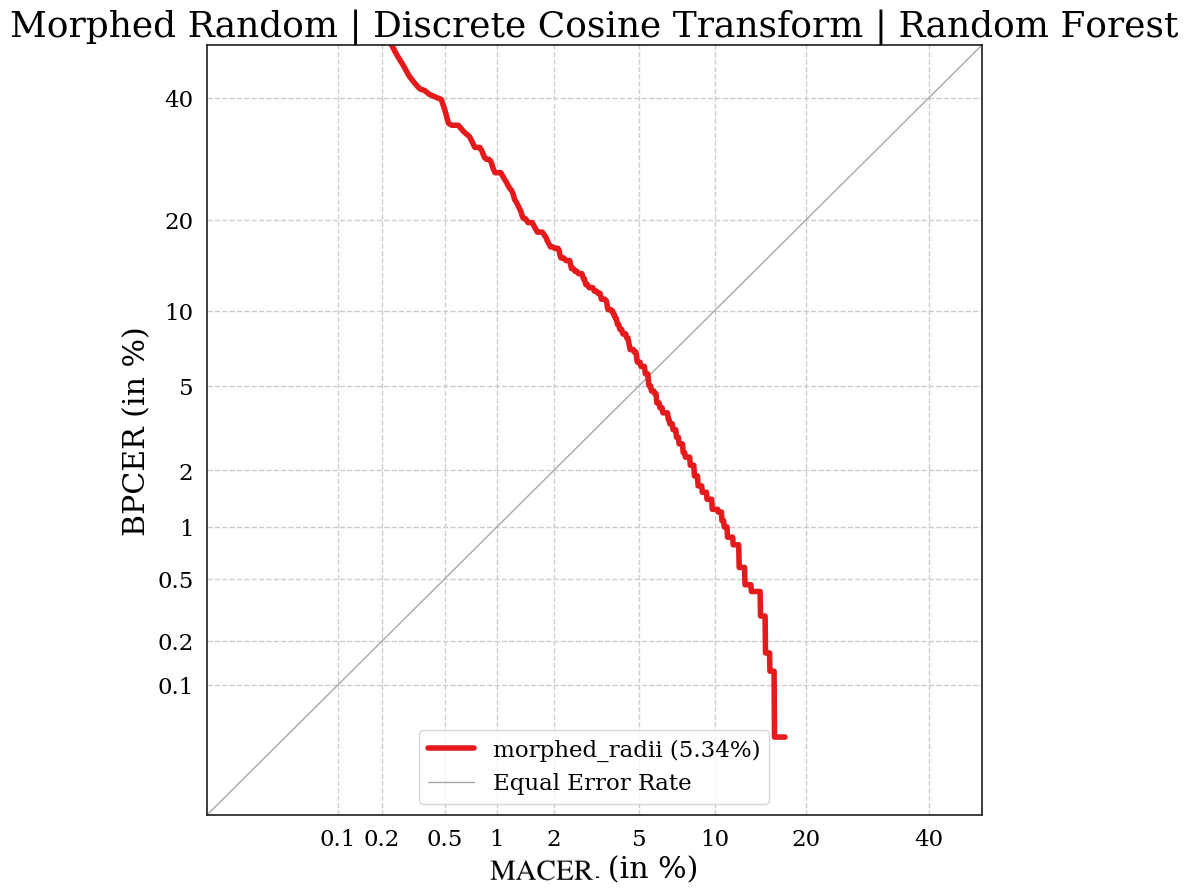}}\\
    \subfloat{\includegraphics[scale=0.15]{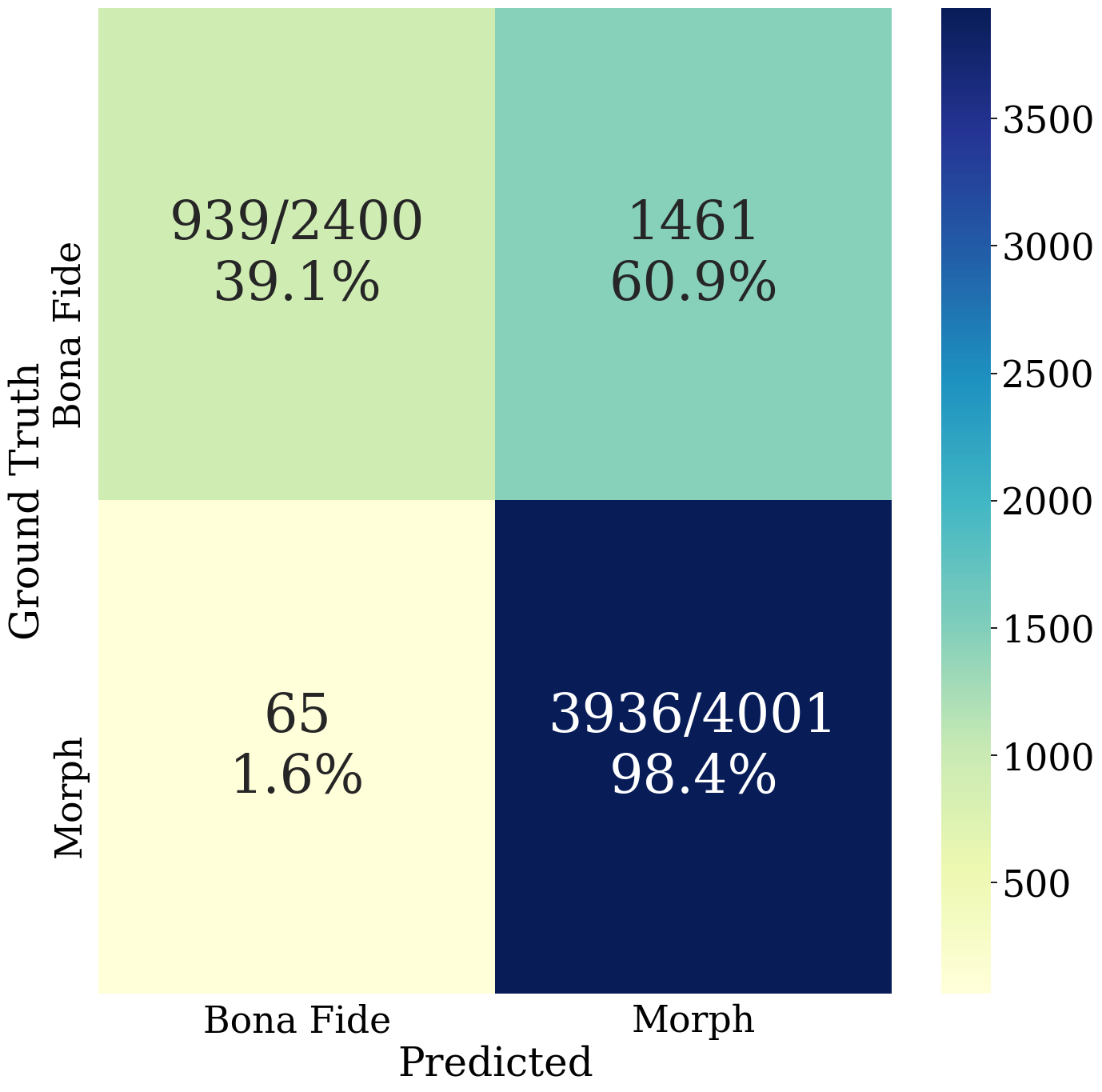}}\hfil%
    \subfloat{\includegraphics[scale=0.15]{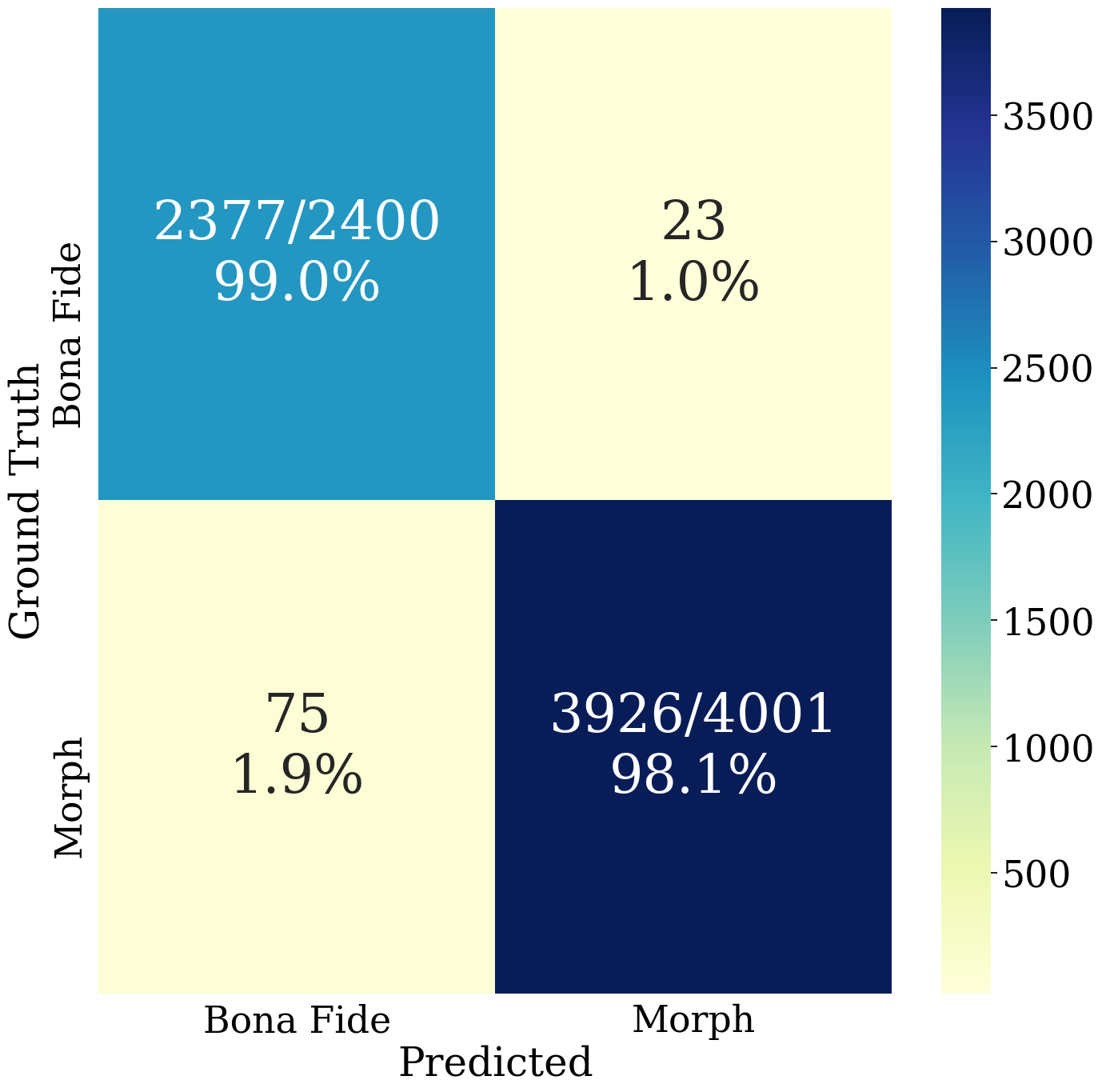}}\hfil%
    \subfloat{\includegraphics[scale=0.24]{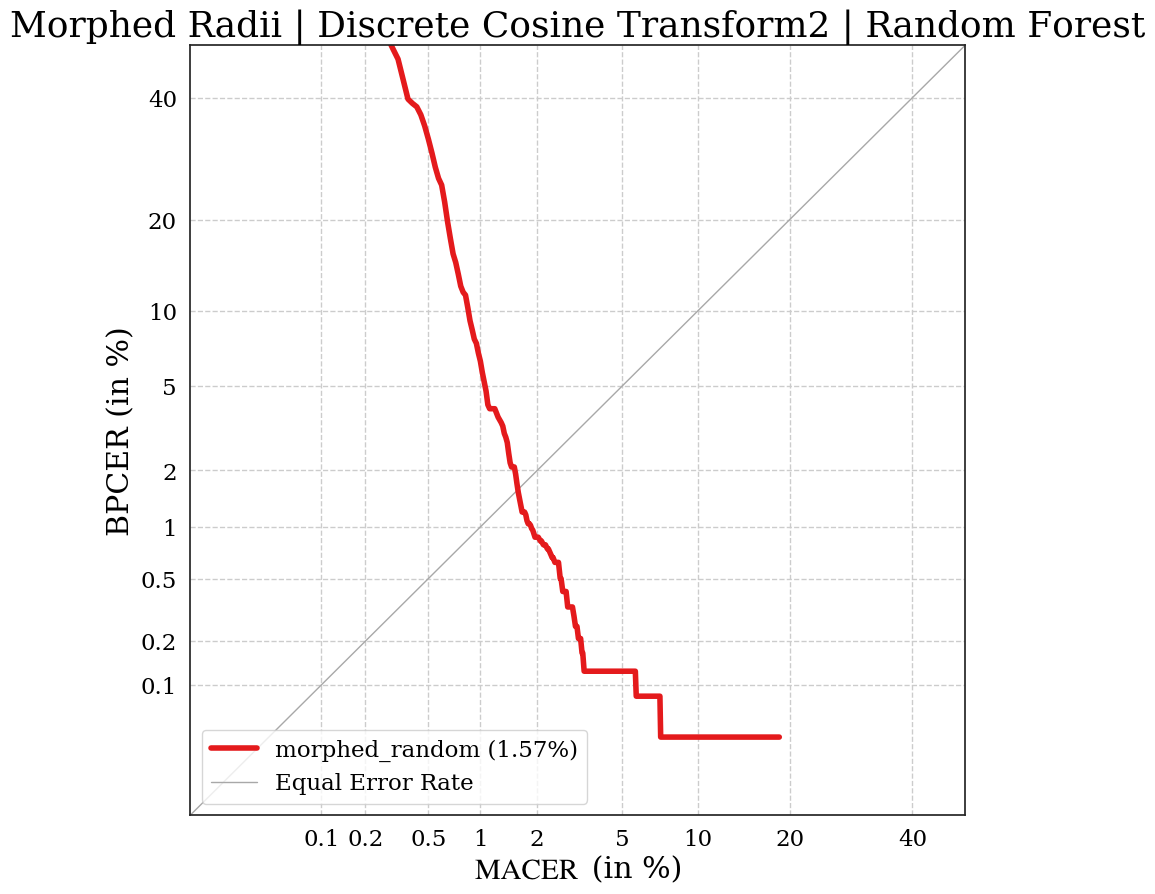}}
    \caption{\emph{Top images: \textbf{random pair} selection}. Top left: Confusion matrix for greyscale morph images used as input to the RF classifier. Top middle: Confusion matrix for DCT morph images used as input to the RF classifier. Top right: DET Curve for DCT morph images used as input to the RF classifier. \emph{Bottom images: \textbf{radius pair} selection}. Bottom left: Confusion matrix for greyscale morph images used as input to the RF classifier. Bottom middle: Confusion matrix for DCT morph images used as input to the RF classifier. Bottom right: DET Curve for DCT morph images used an input to the RF classifier.}
    \label{fig:DET-class}
\end{figure*}

Figure~\ref{fig:DET-class} shows, from left to right, the confusion matrixes using the EER as a threshold for greyscale images and DCT, respectively. Furthermore, the DET curve for bona fide iris image classification versus morphed images generated from random pairs is also presented. The same information is reported in the bottom images but for radius pair selection. We can observe the BPCER\tu{10} of 1.85\% and 0.053\% and a BPCER\tu{100} of 29.85\% and 5.13\%, respectively, for random pairs versus radius pairs selection from the DET curve for each feature extracted.

The confusion matrix shows that DFT achieved better results than directly using the greyscale morphed images.

\section{Conclusions}
\label{sec:conclusions}

This work shows that creating morph images from periocular iris images is feasible and challenging for iris recognition systems. In the state-of-the-art methods, more than (90\%) of morph images can vulnerate the iris recognition systems for both datasets analysed: Notre-Dame-LG4000-LR and CASIA-IrisV4. Creating morph images from periocular images has several advantages because we can obtain the periocular, normalised and iris-code images from the same morph images. 
The S-MAD approach also shows that it is feasible to detect morphed images. The images generated from radius pair selection are more challenging to detect.

The proposed IRS based on the siamese architecture SiamIris is very promising, as it outperforms a state-of-the-art method in the $d'$, FNMR\tu{10} and FNMR\tu{20} metrics. The best configuration uses the periocular image as the input and ResNet50 as the backbone.

As future work, much effort must be made in order to compare iris morphed images with commercial systems and other implementations. While the morph itself is possible, much like the double identity fingerprint and iris, the detection of the morph attack does not seem to exist or be reported as a real attack. We believe that today, this kind of attack would occur as any other presentation attack and would be detected as such.

\section*{Acknowledgements}

This work was supported by the European Union’s Horizon 2020 research and innovation program under grant agreement 883356, and the German Federal Ministry of Education and Research and the Hessian Ministry of Higher Education, Research, Science and the Arts within their joint support of the National Research Center for Applied Cybersecurity ATHENE.

\bibliographystyle{IEEEtran}

\bibliography{bibliography.bib}

\begin{IEEEbiography}[{\includegraphics[width=0.9in,height=1.25in,clip,keepaspectratio]{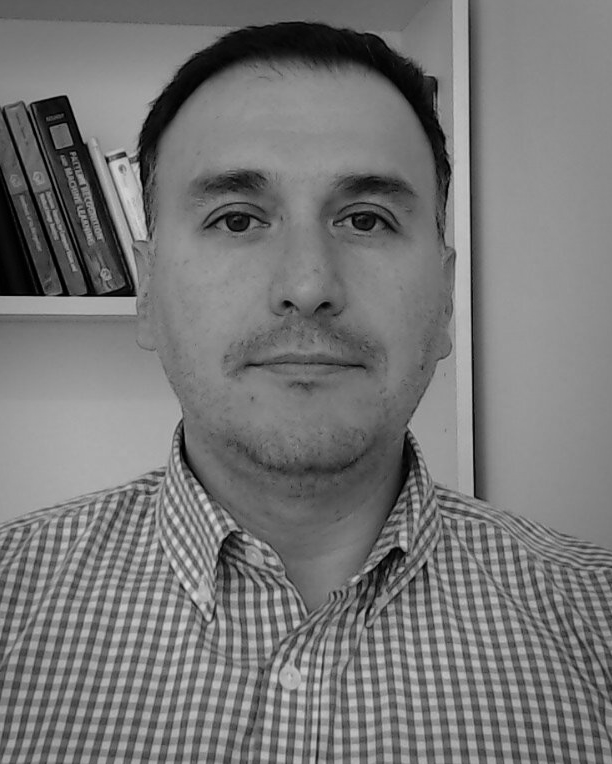}}]{Juan Tapia} received the P.E. degree in electronics engineering from Universidad Mayor, in 2004, and the M.S. and Ph.D. degrees in electrical engineering from the Department of Electrical Engineering, Universidad de Chile, in 2012 and 2016, respectively. In addition, he spent one year of internship with the University of Notre Dame. In 2016, he received the Award for Best Ph.D. Thesis. From 2016 to 2017, he was an Assistant Professor at Universidad Andres Bello. From 2018 to 2020, he was the Research and Development Director for the electricity and electronics area with INACAP, Universidad Tecnologica de Chile, the Research and Development Director of TOC Biometrics Company, and an International Consultor on biometrics for face, iris applications and forensic/tampering ID-card detection. He is currently an Entrepreneur and a Senior Researcher with Hochschule Darmstadt (HDA), leading EU projects, such as iMARS, EINSTEIN and CarMen. His main research interests include pattern recognition and deep learning applied to iris biometrics, morphing, feature fusion, and feature selection. He serves as a reviewer for a number of journals and conferences. He is on behalf of the German DIN Member of the ISO/IEC Sub-Committee 37 on biometrics.
\end{IEEEbiography}
\vspace{-0.3cm}

\begin{IEEEbiography}[{\includegraphics[width=0.9in,height=1.25in,clip,keepaspectratio]{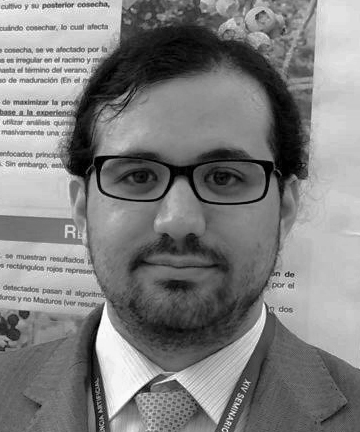}}]{Sebastian Gonzalez} received the B.Sc. degree in Computer Science from Universidad Andres Bello in 2019, and is currently pursuing a M.Sc. degree in Computer Science from Universidad de Santiago de Chile. From 2019 to 2023, he worked as a researcher at TOC Biometrics. He has been involved in different research projects encompassing computer vision, pattern recognition, and deep learning applied to biometrics. His main interests include topics such as presentation attack detection, classification, segmentation, and applied research.
\end{IEEEbiography}
\vspace{-0.3cm}

\begin{IEEEbiography}[{\includegraphics[width=0.9in,height=1.25in,clip,keepaspectratio]{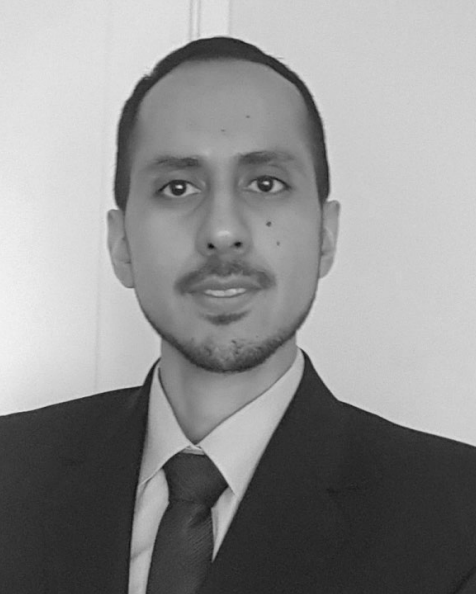}}]%
{Daniel Benalcazar}
(Member, IEEE) was born in Quito, Ecuador, in 1987. He received a B.S. degree in electronics and control engineering from Escuela Politecnica Nacional, Quito, in 2012, an M.S. degree in electrical engineering from The University of Queensland, Australia, in 2014, with a minor in biomedical engineering, and a Ph.D. degree in electrical engineering from the Universidad de Chile, Chile, in 2020. From 2015 to 2016, he worked as a Professor at the Central University of Ecuador. Ever since, he has participated in research projects in biomedical engineering and biometrics. Currently, he is a Researcher at SOVOS, Chile.
\end{IEEEbiography}

\begin{IEEEbiography}[{\includegraphics[width=0.9in,height=1.25in,clip,keepaspectratio]{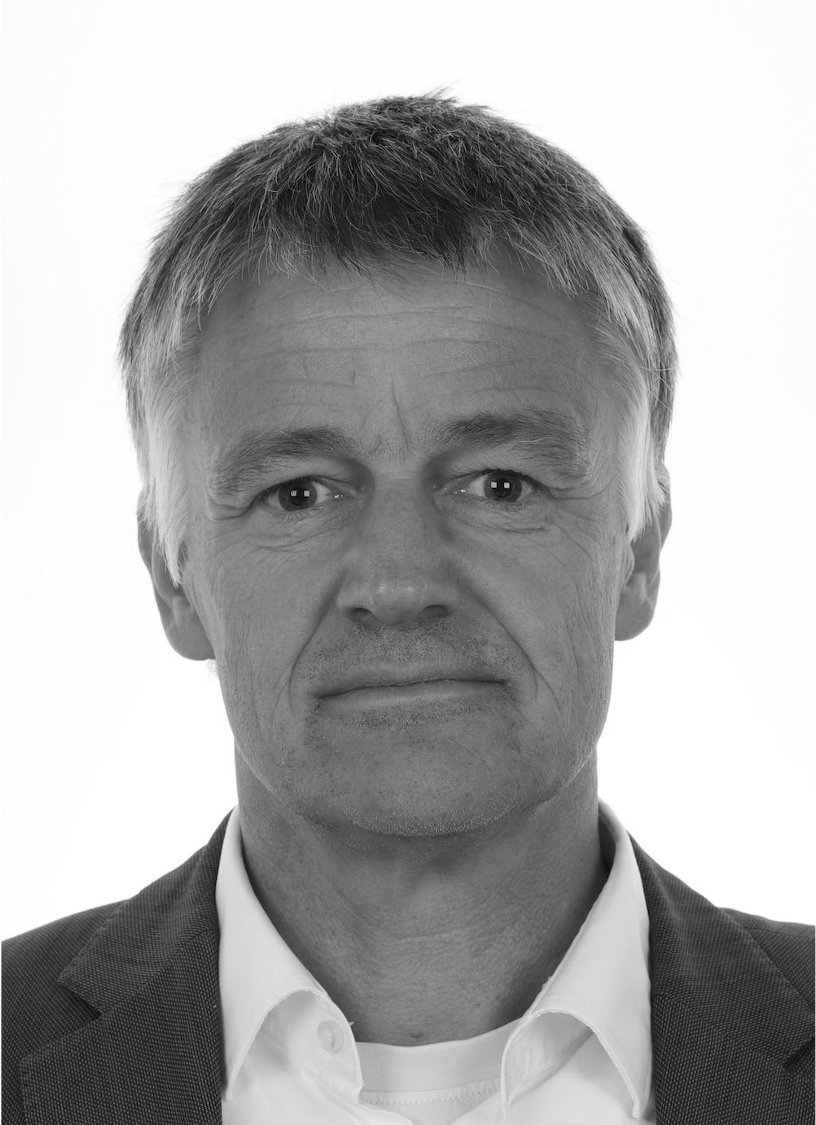}}]{Christoph Busch} is a member of the Department of Information Security and Communication Technology (IIK) at the Norwegian University of Science and Technology (NTNU), Norway. He holds a joint appointment with the computer science faculty at Hochschule Darmstadt (HDA), Germany. Further, he lectures the course Biometric Systems at Denmark’s DTU since 2007. On behalf of the German BSI he has been the coordinator for the project series BioIS, BioFace, BioFinger, BioKeyS Pilot-DB, KBEinweg and NFIQ2.0. In the European research program, he was the initiator of the Integrated Project 3D-Face, FIDELITY and iMARS. Further, he was/is partner in the projects TURBINE, BEST Network, ORIGINS, INGRESS, PIDaaS, SOTAMD, RESPECT and TReSPAsS. He is also principal investigator at the German National Research Center for Applied Cybersecurity (ATHENE). Moreover Christoph Busch is co-founder and member of the board of the European Association for Biometrics (www.eab.org) which was established in 2011 and assembles in the meantime more than 200 institutional members. Christoph co-authored more than 700 technical papers and has been a speaker at international conferences. He is a member of the editorial board of the IET journal.
\end{IEEEbiography}

\end{document}